\documentclass[final,5p,times,twocolumn]{elsarticle}

\usepackage{amssymb}
\usepackage{graphicx}
\usepackage{multirow}
\usepackage{longtable}
\usepackage{rotating}
\usepackage{color}
\usepackage{makecell}
\usepackage{hyperref}
\usepackage{cite}
\usepackage{bm}
\usepackage{lineno}
\graphicspath{{figures/}}

\journal{XXXX}

\begin{document}

\begin{frontmatter}

\title{RFN-Nest: An end-to-end residual fusion network for infrared and visible images}


\author[a]{Hui Li}

\author[a]{Xiao-Jun Wu\corref{correspondingauthor}}

\author[b]{Josef Kittler}

\address[a]{Jiangsu Provincial Engineering Laboratory of Pattern Recognition and Computational Intelligence, \\ School of Artificial Intelligence and Computer Science, Jiangnan University, \\ 214122, Wuxi, China \fnref{address1}}
\address[b]{The Center for Vision, Speech and Signal Processing, University of Surrey, Guildford, GU2 7XH, UK. \fnref{address2}}

\cortext[correspondingauthor]{Corresponding author email: wu\_xiaojun@jiangnan.edu.cn}


\begin{abstract}
In the image fusion field, the design of deep learning-based fusion methods is far from routine. It is invariably fusion-task specific and requires a careful consideration. The most difficult part of the design is to choose an appropriate strategy to generate the fused image for a specific task in hand. Thus, devising learnable fusion strategy is a very challenging problem in the community of image fusion. To address this problem, a novel end-to-end fusion network architecture (RFN-Nest) is developed for infrared and visible image fusion. We propose a residual fusion network (RFN) which is based on  a residual architecture to replace the traditional fusion approach. A novel detail-preserving loss function, and a feature enhancing loss function are proposed to train RFN. The fusion model learning is accomplished by a novel two-stage training strategy. In the first stage, we train an auto-encoder based on an innovative nest connection (Nest) concept. Next, the RFN is trained using the proposed loss functions. The experimental results on public domain data sets show that, compared with the existing methods, our end-to-end fusion network delivers a better performance  than the state-of-the-art methods in both subjective and objective evaluation. The code of our fusion method is available at \url{https://github.com/hli1221/imagefusion-rfn-nest}.
\end{abstract}


\end{frontmatter}


\section{Introduction}

Due to the physical limitations of imaging sensors, it is very difficult to capture an image of a scene that is of uniformly good quality. Image fusion  plays an important role in this context. Its aim is to reconstruct a perfect image of the scene from multiple samples that provide complementary information about the visual content. It has  many applications, such as object tracking \citep{li2019rgb} \citep{li2018learning} \citep{luo2019thermal}, self-driving and video surveillance \citep{shrinidhi2018ir}. The fusion task requires algorithms to generate a single image which amalgamates the  complementary information conveyed by different source images \citep{ma2019infrared}\citep{li2017pixel}\citep{liu2018deep}.

Image fusion involves three key processes: feature extraction, fusion strategy and reconstruction. Most of the existing fusion research focuses on one or more of these elements to improve the fusion performance. The existing fusion methods can be classified into two categories: traditional algorithms and deep learning-based methods. In the traditional algorithm category, multi-scale transform methods  \citep{pajares2004wavelet}\citep{ben2005multiscale}\citep{yang2010image}\citep{li2013image} are widely applied to extract multi-scale features from the source images. The feature channels are combined by an appropriate fusion strategy. Finally, the fused image is reconstructed by an inverse multiscale transform. Obviously, the fusion performance of these algorithms is highly dependent on the feature extraction method used.

Following this direction, sparse representation (SR) \citep{wright2008robust} and low-rank representation (LRR) \citep{liu2010robust}\citep{liu2012robust} have been applied to extract salient features from the source images. In SR and LRR based fusion methods \citep{zhang2013dictionary}\citep{liu2017infrared}\citep{gao2017image}\citep{li2017multi}, the sliding window technique is used to decompose source images into image patches. A matrix is constructed using these image patches, in which each column is a reshaped image patch. This matrix is fed into SR (or LRR) to calculate SR (or LRR) coefficients which are considered  as image features. By virtue of this operation, the image fusion problem is transformed to the one of coefficient fusion. The fused coefficients are generated by an  appropriate fusion strategy and used to reconstruct the fused image in the SR (or LRR) framework. Beside the above, other SR based approaches and other signal processing methods \citep{lu2014infrared}\citep{yin2017novel} have been suggested in the literature.

Although the traditional fusion methods have achieved good fusion performance, they have drawbacks: (1) The fusion performance highly depends on handcrafted features \citep{zhang2013dictionary}\citep{li2017multi}\citep{liu2017jsr}, as it is difficult to find a universal feature extraction method for different fusion tasks; (2) Different fusion strategies may be required to work with different features; (3) For SR and LRR based methods, the dictionary learning is very time-consuming; (4) Complex source images pose a challenge for SR (or LRR) based fusion methods.

To overcome these drawbacks, deep learning based fusion methods have been developed, which can be grouped into three categories according to the three key elements of the fusion process: deep feature extraction, fusion strategy and end-to-end training. In the feature extraction direction, deep learning methods are utilized to extract deep representation of the  information conveyed by the source images \citep{li2018infrared}\citep{li2019infrared}\citep{song2018multi}\citep{li2018densefuse}\citep{li2020mdlatlrr}. Different fusion strategies have been suggested to reconstruct the fused image. In other fusion methods \citep{liu2016image}\citep{liu2017multi}, deep learning is also used to design the fusion strategy. In \citep{liu2016image}\citep{liu2017multi}, convolutional sparse representation and convolutional neural network are utilized to generate a decision map for the source images. Using the learned decision map, the fused images are obtained by appropriate post-processing. Although these fusion methods achieve good fusion performance, the fusion strategy and the post-processing are tricky to design. To avoid the limitations of handcrafted solutions, some end-to-end fusion frameworks were presented (FusionGAN \citep{ma2019fusiongan}, FusionGANv2 \citep{ma2020infrared}, DDcGAN \citep{ma2020ddcgan}). These frameworks are based on adversarial learning which avoids the shortcoming of handcrafted features and fusion strategies. However, even the state of the art methods, FusionGANv2 \citep{ma2020infrared} and DDcGAN \citep{ma2020ddcgan}, face challenges to preserve image detail adequately. To preserve more detail background information from visible images, a nest connection based autoencoder fusion network (NestFuse \citep{li2020nestfuse}) was proposed. Although NestFuse obtains good performance in detail information preservation, the fusion strategy is still not learnable.

To address these problems, in this paper, we propose a novel end-to-end fusion framework (RFN-Nest). Our network contains three parts: an encoder network, residual fusion network (RFN) which is designed to extract fused multi-scale deep features, and a decoder network based on nest connection \citep{zhou2018unet++}. Although the encoder and decoder architecture of the proposed network is similar to the NestFuse \citep{li2020nestfuse}, the fusion strategy, the training strategy and the loss function are totally different. 

Firstly, instead of fusing handcrafted features in NestFuse \citep{li2020nestfuse}, several simple yet efficient learnable fusion networks (RFN) have been designed and inserted into the autoencoder architecture. With the RFN, the autoencoder-based fusion network is upgraded to an end-to-end fusion network. Secondly, as RFN is a learnable structure, it is important that the encoder and decoder exhibit powerful feature extraction and feature reconstruction abilities, respectively. Thus, we develop a two-stage training strategy to train our fusion network (encoder, decoder and RFN networks). Thirdly, to train the proposed RFN networks, we design a new loss function ($L_{RFN}$) to preserve the detail information from visible image and maintain the salient features from infrared image, simultaneously.

The main contributions of RFN-Nest can be summarized as follows,

(1)	A novel residual fusion network(RFN) is proposed to supersede handcrafted fusion strategies. Although many methods \citep{li2018infrared}\citep{song2018multi}\citep{ram2017deepfuse}\citep{li2018densefuse} now use deep features to achieve good performance, the heuristic approach to selecting a suitable fusion strategy is  their weakness. The proposed RFN is a learnable fusion network that overcomes this weakness.

(2)	A two-stage training strategy is developed to design our network. The feature extraction and feature reconstruction abilities are the key for the encoder and decoder networks. Using only one stage training strategy to simultaneously train the whole network (encoder, decoder and RFN networks) is insufficient. Inspired by \citep{li2018densefuse}, firstly, the encoder and the decoder network are trained as an auto-encoder. With the fixed encoder and decoder, the RFN networks are trained using an appropriate loss function.


(3)	A loss function capable of preserving the image detail, together with a feature enhancing loss function are designed to train our RFN networks. We show that with these loss functions, more detail information and image salient features are preserved in the fused image.

(4)	We show that, compared with the state-of-the-art fusion methods, the proposed RFN-Nest framework exhibits better fusion performance on  public datasets in both subjective visual assessment and objective evaluation.

The rest of our paper is structured as follows. In Section \ref{sec-relate}, we briefly review the related work on deep learning-based fusion. The proposed fusion framework is described in detail in Section \ref{sec-proposed}. The experimental results are presented in Section \ref{sec-experiments} and Section \ref{rgbt-tracking}. Finally, we draw the paper to conclusion in Section \ref{sec-con}.

\section{Related Works}
\label{sec-relate}
Recently, many deep learning methods have been developed for image fusion. Most of them are based on convolutional neural networks (CNN). These methods can be classified into the non end-to-end learning  and end-to-end learning categories. In this section, we briefly overview the most representative deep learning based methods from these two categories.

\subsection{Non End-to-end Methods}
In the early days, deep learning neural networks were used to extract deep features as a bank of ``decision" maps \citep{li2018infrared}\citep{li2019infrared}\citep{song2018multi}. In \citep{li2018infrared}, Li et al. proposed a fusion framework based on a pre-trained network (VGG-19 \citep{simonyan2014very}). Firstly, the source images are decomposed into salient parts (texture and edges) and base parts (contour and luminance). Then, VGG-19 is used to extract multi-level deep features from the salient parts. At each level, the decision maps are computed from the deep features and a candidate fused salient part is generated. The fused image is reconstructed by combining the fused base parts and the fused salient parts using an appropriate fusion strategy. In \citep{li2019infrared}, the pre-trained ResNet-50 \citep{he2016deep} is utilized to extract deep features from the source images directly. A decision map is obtained by zero-phase component analysis(ZCA) and $l_1$-norm. The PCANet-based fusion method \citep{song2018multi} also follows this framework to generate the fused image, in which PCANet, instead of VGG-19 or ResNet-50,  is used to extract the features.

In addition to pure feature extraction, in \citep{liu2016image}\citep{liu2017multi}, the two key processes (feature extraction and fusion strategy) are implemented by a single network. In \citep{liu2017multi}, a decision map is generated by a CNN trained on image patches of multiple blurred versions of the input image. In \citep{liu2016image}, the convolutional sparse representation instead of CNN is utilized to extract features and to generate a decision map. From the generated decision map, the fused image can easily be reconstructed.

\begin{figure*}[ht]
	\centering
	\includegraphics[width=0.8\linewidth]{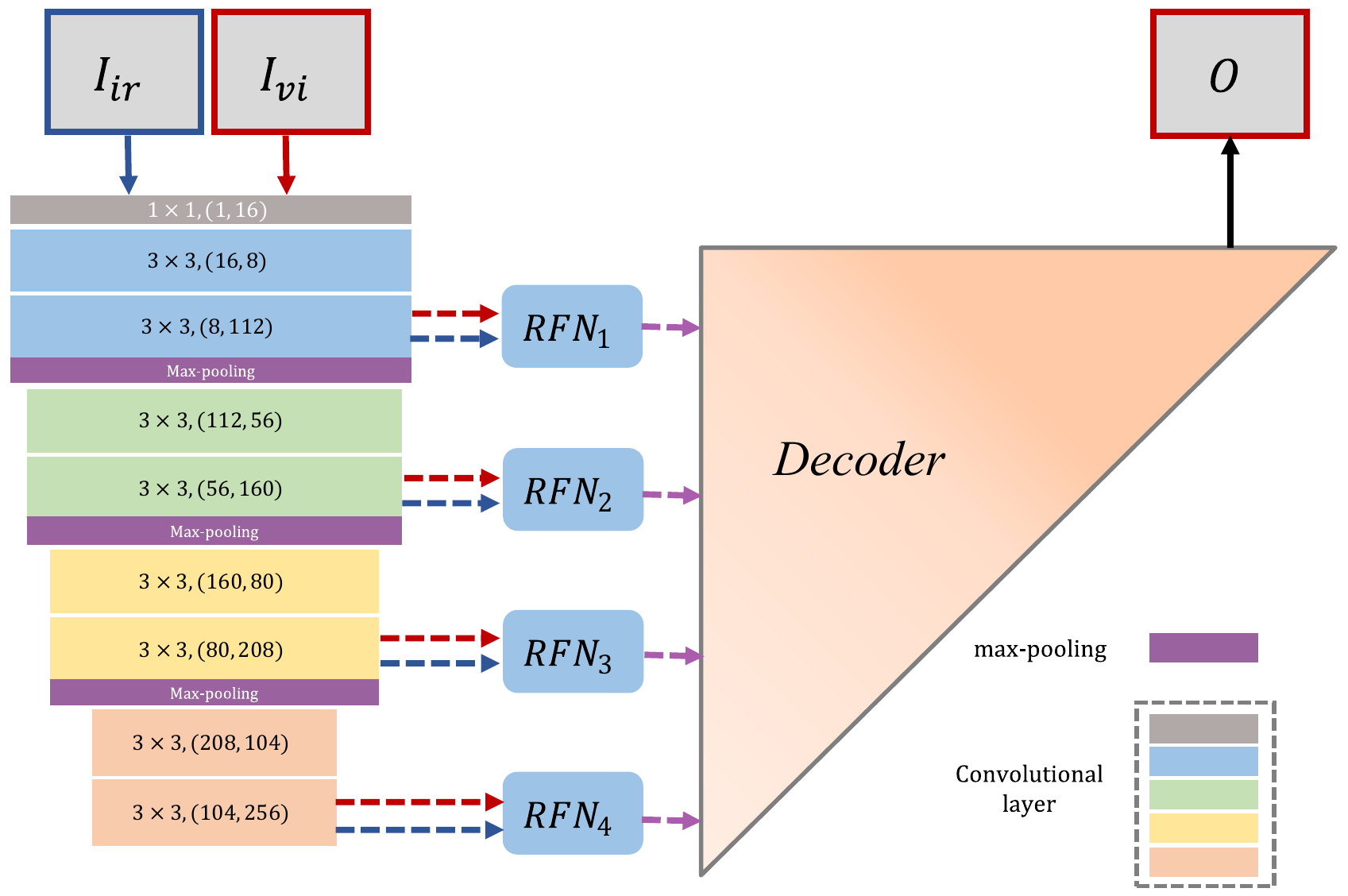}
	\caption{The framework of RFN-Nest. `$RFN_{1-4}$' denote the residual fusion network. The nest connection-based decoder network(`Decoder') will be introduced later. '$3 \times 3, (16, 8)$' means the kernel size is $3 \times 3$, input channel is 16 and output channel is 8 in a convolutional layer.}
	\label{fig:framework}
\end{figure*}

Besides the above methods, a deep auto-encoder network based fusion framework was proposed in \citep{li2018densefuse}. Inspired by DeepFuse \citep{ram2017deepfuse}, the authors proposed a novel network architecture which contains an encoder, a fusion layer and a decoder. The dense block \citep{huang2017densely} based encoder network was adopted as it extracts more complementary deep features from the source images. 
In their framework, the fusion strategy becomes very important. 

Inspired by DenseFuse\citep{li2018densefuse} and the architecture in \citep{zhou2018unet++}, Li et al. proposed NestFuse\citep{li2020nestfuse} to preserve more detail background information from visible images, while enhancing the salient features in infrared images. Additionally, a novel spatial/channel attention models are designed to fuse the multi-scale deep features. Although these frameworks achieve good fusion performance, it is very difficult to find an effective handcrafted fusion strategy for image fusion.

\subsection{End-to-end Methods}
To eliminate the arbitrariness of handcrafted features and fusion strategies, several end-to-end fusion frameworks have been suggested \citep{ma2019fusiongan} \citep{ma2020infrared} \citep{zhang2020ifcnn} \citep{zhang2020PMGI} \citep{ma2020ddcgan} \citep{xu2020u2fusion}.

In \citep{ma2019fusiongan}, a GAN-based fusion framework (FusionGAN) was introduced to the infrared and visible image fusion field. The generator network is the engine which computes the fused image, while the discriminator network constrains the fused image to contain the detail information from the visible image. The loss function has two terms: content loss and discriminator loss. Due to the content loss, the fused image tends to become similar to the infrared image, failing to preserve the image detail, in spite of the discriminator network.

To preserve more detail information from the visible images, the authors of \citep{ma2020infrared} proposed a new version of FusionGAN which was named FusionGANv2. In this new version, the authors deepen the generator and discriminator networks, endowing them with more powerful feature representation ability. In addition, two new loss functions, namely detail loss and target edge-enhancement loss,  were presented to preserve the detail information. With these improvements, the fused images reconstructed more scene details, with clearly highlighted edge-sharpened targets.

A general end-to-end image fusion network (IFCNN) \citep{zhang2020ifcnn} was also proposed, which is a simple yet effective fusion method. In IFCNN, two convolutional layers are utilized to extract deep features from the source images. Element-wise fusion rules (elementwise-maximum, elementwise-sum, elementwise-mean) are used to fuse the deep features. The fused image is generated from the fused deep features by two convolutional layers. Although IFCNN achieves a satisfactory fusion performance in multiple image fusion tasks, its architecture is too simplistic to extract powerful deep features, and the fusion strategies designed using a traditional way are not optimal.

\section{The Proposed Fusion Framework}
\label{sec-proposed}
The proposed fusion network is introduced in this section. Firstly, the architecture of our network is presented in Section \ref{afn}. The advocated two-stage training strategy is described in Section \ref{strategy}. 


\subsection{The Architecture of the Fusion Network}
\label{afn}
The RFN-Nest is an end-to-end fusion network, the architecture of which is shown in Fig.\ref{fig:framework}. RFN-Nest contains three parts: encoder (left part), residual fusion network ($RFN_{1-4}$) and decoder (right part). For a convolutional layer, ``$k \times k, (\bm{in}, \bm{out})$'' means the kernel size is $k \times k$, input channel is $\bm{in}$ and output channel is $\bm{out}$.

With the max pooling operation in the encoder network, multi-scale deep features can be extracted from the source images. The RFN is utilized to fuse multi-modal deep features extracted at each scale. While the shallow layer features preserve more detail information, the deeper layer features convey semantic information, which is important for reconstructing the salient features. Finally, the fused image is reconstructed by the nest connection-based decoder network, which fully exploits the multi-scale structure of the features.

As shown in Fig.\ref{fig:framework}, $I_{ir}$ and $I_{vi}$ indicate the source images (infrared image and visible image). $O$ denotes the output of RFN-Nest, that is the fused image. ``$RFN_m$'' means one residual fusion network for  deep features at scale $m$. The architecture of the encoder in our framework is constituted by four RFN networks,  $m\in{\{1,2,3,4\}}$. These RFN networks share the same architecture but with different weights.

We now introduce the RFN and the decoder in detail.

\subsubsection{Residual Fusion Network (RFN)}
\begin{figure}[ht]
	\centering
	\includegraphics[width=0.85\linewidth]{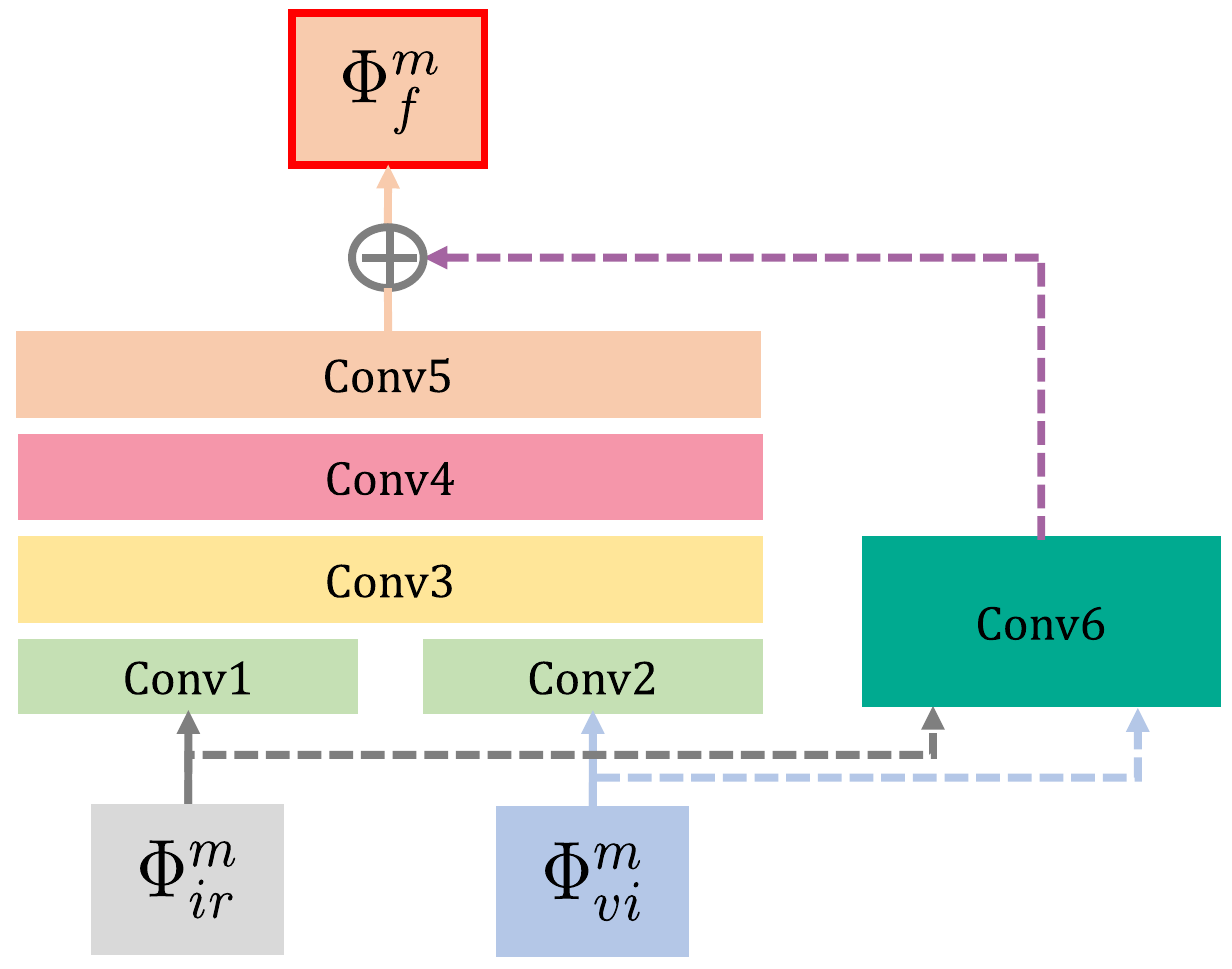}
	\caption{The architecture of $RFN_m$.}
	\label{fig:rfn}
\end{figure}

The RFN is based on the concept of residual block \citep{he2016deep} which has been adapted to the task of image fusion. The RFN architecture is shown in Fig.\ref{fig:rfn}.

In Fig.\ref{fig:rfn}, $\Phi_{ir}^m$ and $\Phi_{vi}^m$ indicate the $m$-th scale deep features extracted by the encoder network, with $m\in{\{1,2,3,4\}}$ indicating the index of the RFN network. ``Conv1-6'' denote six convolutional layers in RFN. In this residual architecture, the outputs of ``Conv1'' and ``Conv2'' are concatenated as the input of ``Conv3''. ``Conv6'' is the first fusion layer to generate initial fused features. With this architecture, RFN can easily be optimized by our training strategy. The convolutional operations produce the fused deep features $\Phi_f^m$, which are fed into the decoder network.

Thanks to the multi-scale deep features and the proposed learning process, both the image detail and salient structures are preserved by the shallow RFN networks and deep RFN networks, respectively.

\subsubsection{Decoder Network}
\begin{figure}[!ht]
	\centering
	\includegraphics[width=0.85\linewidth]{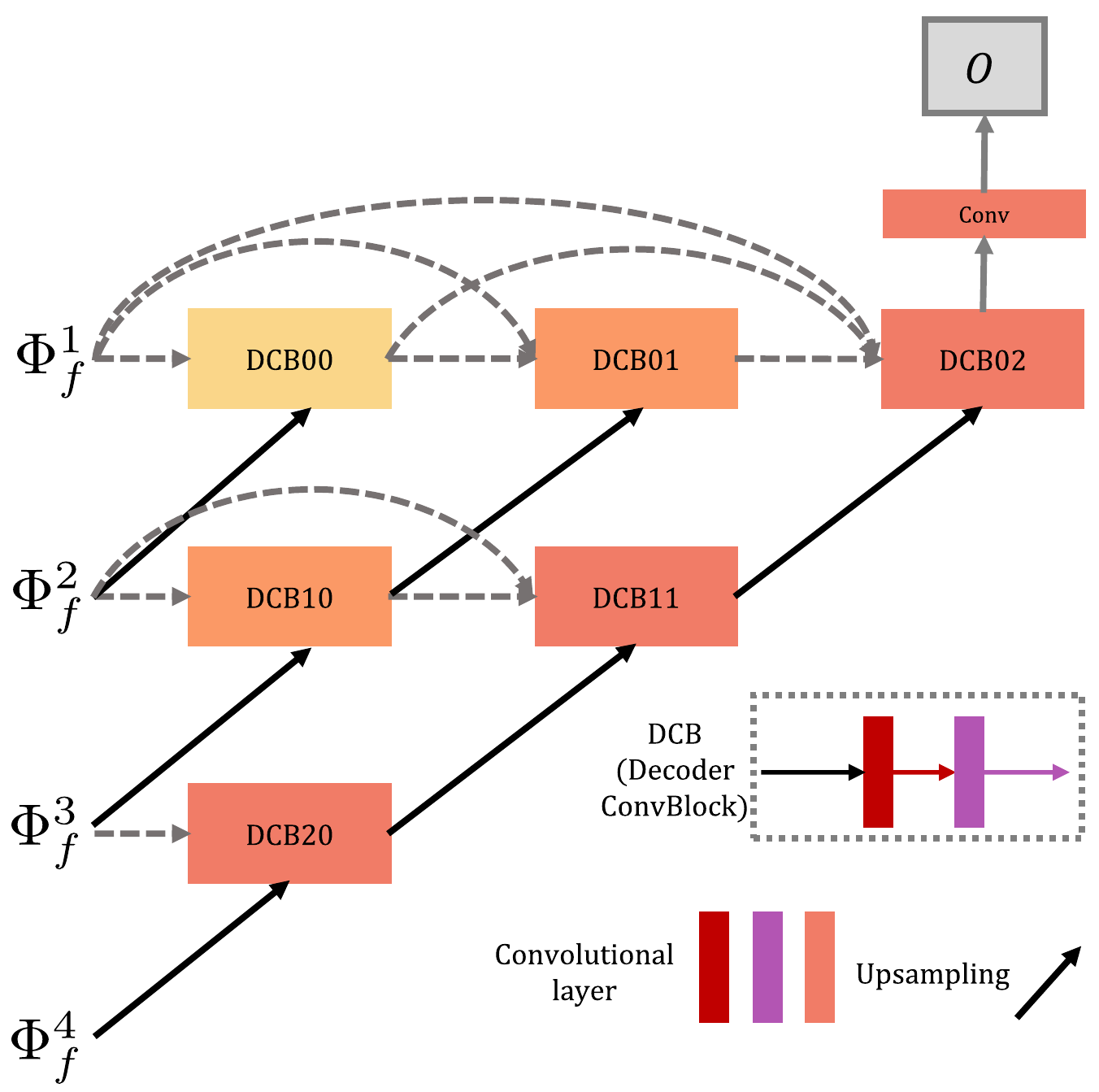}
	\caption{The architecture of the decoder.}
	\label{fig:decoder}
\end{figure}

The decoder network based on the nest connection architecture is shown in Fig.\ref{fig:decoder}. Compared to UNet++ \citep{zhou2018unet++}, regarding the image fusion task, we simplify the network architecture to make it light yet effective to reconstruct fused images, this architecture was also utilized in \citep{li2020nestfuse}.

$\Phi_f^m$ ($m\in{\{1,2,3,4\}}$) denote the fused multi-scale features obtained by the RFN networks. ``DCB'' indicates a decoder convolutional block, which has two convolutional layers. In each row, these blocks are connected by short connections which are similar to the dense block architecture \citep{huang2017densely}. The cross-layer links connect multi-scale deep features in the decoder network.

The output of the network is the fused image reconstructed from the fused multi-scale features.

\subsection{Two-stage Training Strategy}
\label{strategy}

Note that the ability of the encoder in our network to perform feature extraction and that of the decoder to conduct feature reconstruction are absolutely crucial for successful operation. Accordingly, we develop a two-stage training strategy to make sure that each part in our network can achieve the expected performance.

Firstly, the encoder and the decoder are trained as an auto-encoder network to reconstruct the input image. After learning the encoder and decoder networks, in the second training stage, several RFN networks are trained to fuse the multi-scale deep features.

In this section, a novel two-stage training strategy is introduced in detail.

\subsubsection{Training of the Auto-encoder Network}
\begin{figure}[ht]
\centering
\includegraphics[width=0.8\linewidth]{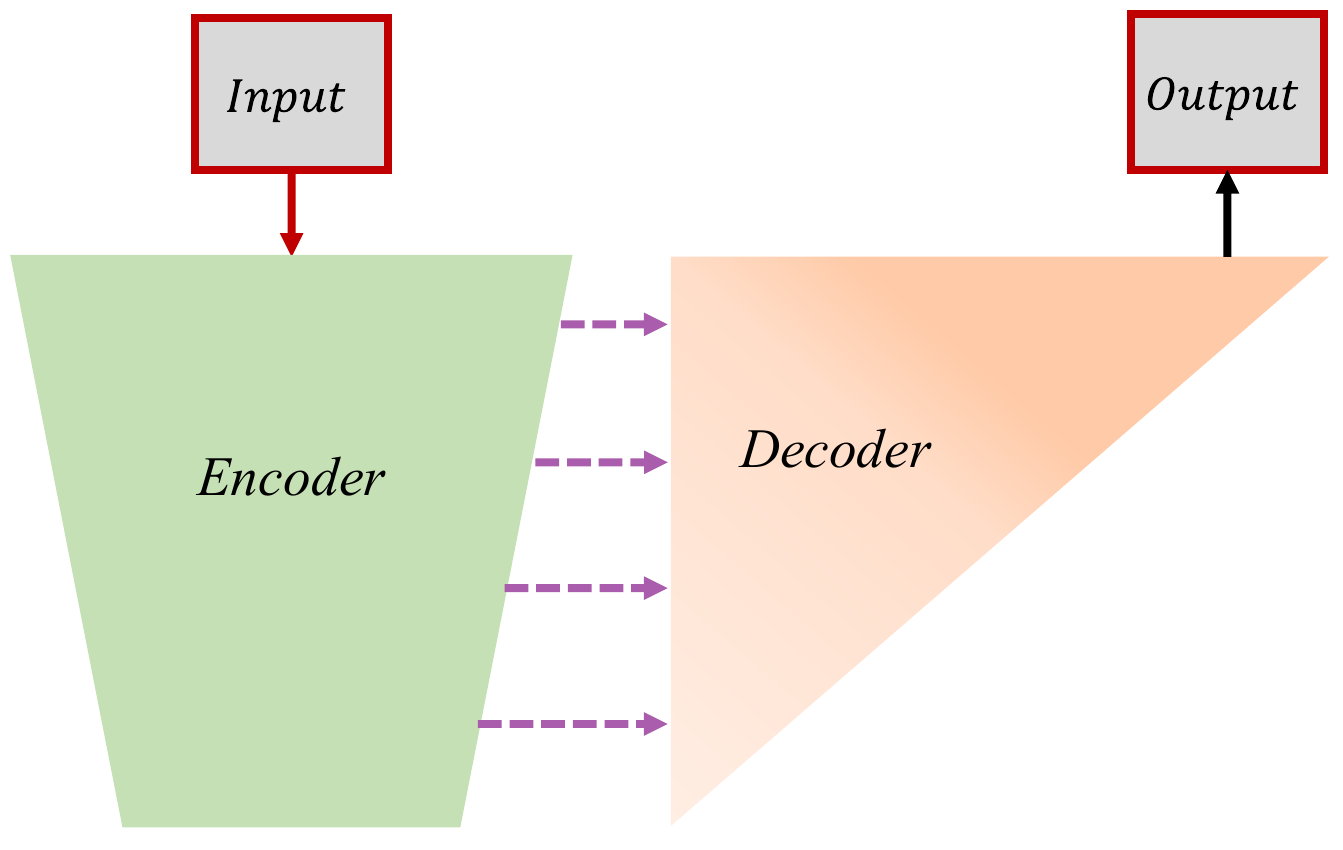}
\caption{The training of the  auto-encoder network.}
\label{fig:train-auto}
\end{figure}

Inspired by DenseFuse \citep{li2018densefuse}, in the first stage, the encoder network is trained to extract multi-scale deep features. The decoder network is trained to reconstruct the input image with multi-scale deep features. The auto-encoder network training framework is shown in Fig.\ref{fig:train-auto}.

In Fig.\ref{fig:train-auto}, $Input$ and $Output$ denote the input image and the output image (both indicate one image), respectively. In contrast to \citep{ram2017deepfuse}\citep{li2018densefuse}\citep{ma2019fusiongan}\citep{ma2020infrared}, our feature extraction part includes a down sampling operation (max pooling), which extracts deep features at four scales. These multi-scale deep features are fed into the decoder network to reconstruct the input image. With short cross-layer connections, the multi-scale deep features are fully used to reconstruct the input image.

The auto-encoder network is trained using the loss function, $L_{auto}$ defined as follows,
\begin{eqnarray}\label{equ:loss-auto}
  	L_{auto}=L_{pixel}+\lambda L_{ssim}
\end{eqnarray}
where $L_{pixel}$ and $L_{ssim}$ denote the pixel loss and the structure similarity (SSIM) loss between the input image ($Input$) and the output image ($Output$). $\lambda$ is the trade-off parameter between $L_{pixel}$ and $L_{ssim}$.

The pixel loss ($L_{pixel}$) is calculated by Eq.\ref{equ:loss-pixel},
\begin{eqnarray}\label{equ:loss-pixel}
  	L_{pixel}=||Output - Input||_F^2
\end{eqnarray}
where the $||\cdot||_F$ is the Frobenius norm. $L_{pixel}$ constrains the reconstructed image to be like the input image at the pixel level.

The SSIM loss ($L_{ssim}$) is defined as,
\begin{eqnarray}\label{equ:loss-ssim}
  	L_{ssim}=1 - SSIM(Output, Input)
\end{eqnarray}
where $SSIM(\cdot)$\footnote{The definition of $SSIM(\cdot)$ is introduced in our supplementary material (Section 1).} is the structural similarity measure \citep{wang2004image} which quantifies the structural similarity of the two images. The structural similarity between $Input$ and $Output$ is constrained by $L_{ssim}$.

\subsubsection{Training of the RFN}
\label{sec-train-rfn}

\begin{figure}[ht]
	\centering
	\includegraphics[width=\linewidth]{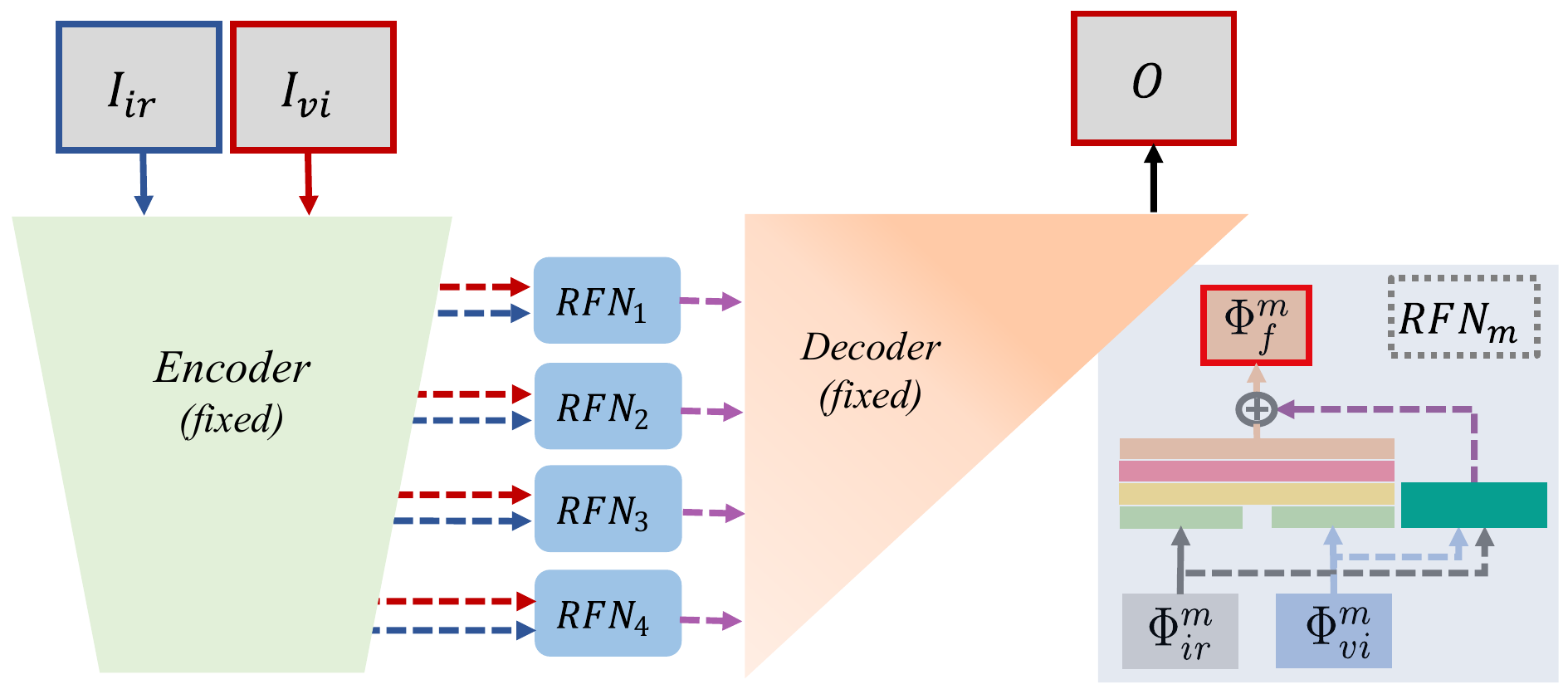}
	\caption{The training of RFN. In the RFN, inputs $\Phi_{ir}^m$ and $\Phi_{vi}^m$ denote infrared and visible deep features, respectively. $\Phi_f^m$ represents the fused deep features obtained by $RFN_m$. In our framework, $m\in{\{1,2,3,4\}}$ indicates the scale of deep features.}
	\label{fig:train-rfn}
\end{figure}

The RFN is proposed to implement a fully learnable fusion strategy. In the second stage, with the encoder and decoder fixed, the RFN is trained with an appropriate loss function. The training process is shown in Fig.\ref{fig:train-rfn}.

The fixed encoder network is utilized to extract multi-scale deep features ($\Phi_{ir}^m$ and $\Phi_{vi}^m$) from the source images. For each scale, an RFN is used to fuse these deep features. Then, the fused multi-scale features ($\Phi_f^m$) are fed into the fixed decoder network.

To train our RFN, we propose a novel loss function $L_{RFN}$, which is defined as,
\begin{eqnarray}\label{equ:loss-rfn}
  	L_{RFN}=\alpha L_{detail} + L_{feature}
\end{eqnarray}
where $L_{detail}$ and $L_{feature}$ indicate the background detail preservation loss function and the target feature enhancement loss function, respectively. $\alpha$ is a trade-off parameter.

In the case of infrared and visible image fusion, most of the background detail information comes from the visible image. $L_{detail}$ aims to preserve the detail information and structural features from visible image, which is defined as
\begin{eqnarray}\label{equ:loss-detail}
  	L_{detail}=1 - SSIM(O, I_{vi})
\end{eqnarray}

As the infrared image contains more salient target features than the visible image, the loss function $L_{feature}$ is designed to constrain the fused deep features so as  to preserve the salient structures. The $L_{feature}$ is defined as,
\begin{eqnarray}\label{equ:loss-feature}
  	L_{feature}=\sum_{m=1}^M w_1(m)||\Phi_f^m - (w_{vi}\Phi_{vi}^m+w_{ir}\Phi_{ir}^m)||_F^2
\end{eqnarray}

In Eq.\ref{equ:loss-feature}, $M$ is the number of the multi-scale deep features, which is set to 4. Owing to the magnitude difference between the scales, $w_1$ is a trade-off parameter vector for balancing the loss magnitudes. It assumes four values $\{1,10,100,1000\}$. $w_{vi}$ and $w_{ir}$ control the relative influence of the visible and infrared features in the fused feature map $\Phi_f^m$. 

As the visible information is constrained by $L_{detail}$ and the aim of $L_{feature}$ is to preserve salient features from the infrared image, in Eq.\ref{equ:loss-feature}, $w_{ir}$ is usually greater than $w_{vi}$.


\begin{figure*}[ht]
	\centering
	\includegraphics[width=0.85\linewidth]{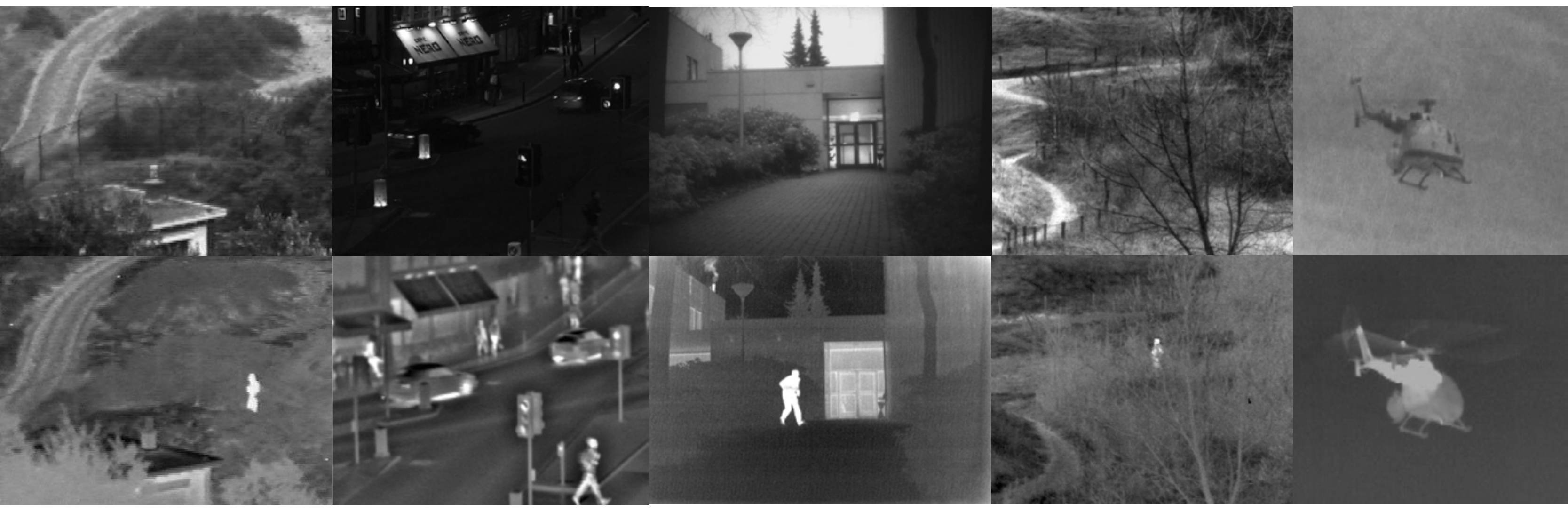}
	\caption{Five pairs of source images. The top row contains visible images, and the second row contains infrared images.}
	\label{fig:example}
\end{figure*}

\begin{figure*}[!ht]
	\centering
	\includegraphics[width=0.85\linewidth]{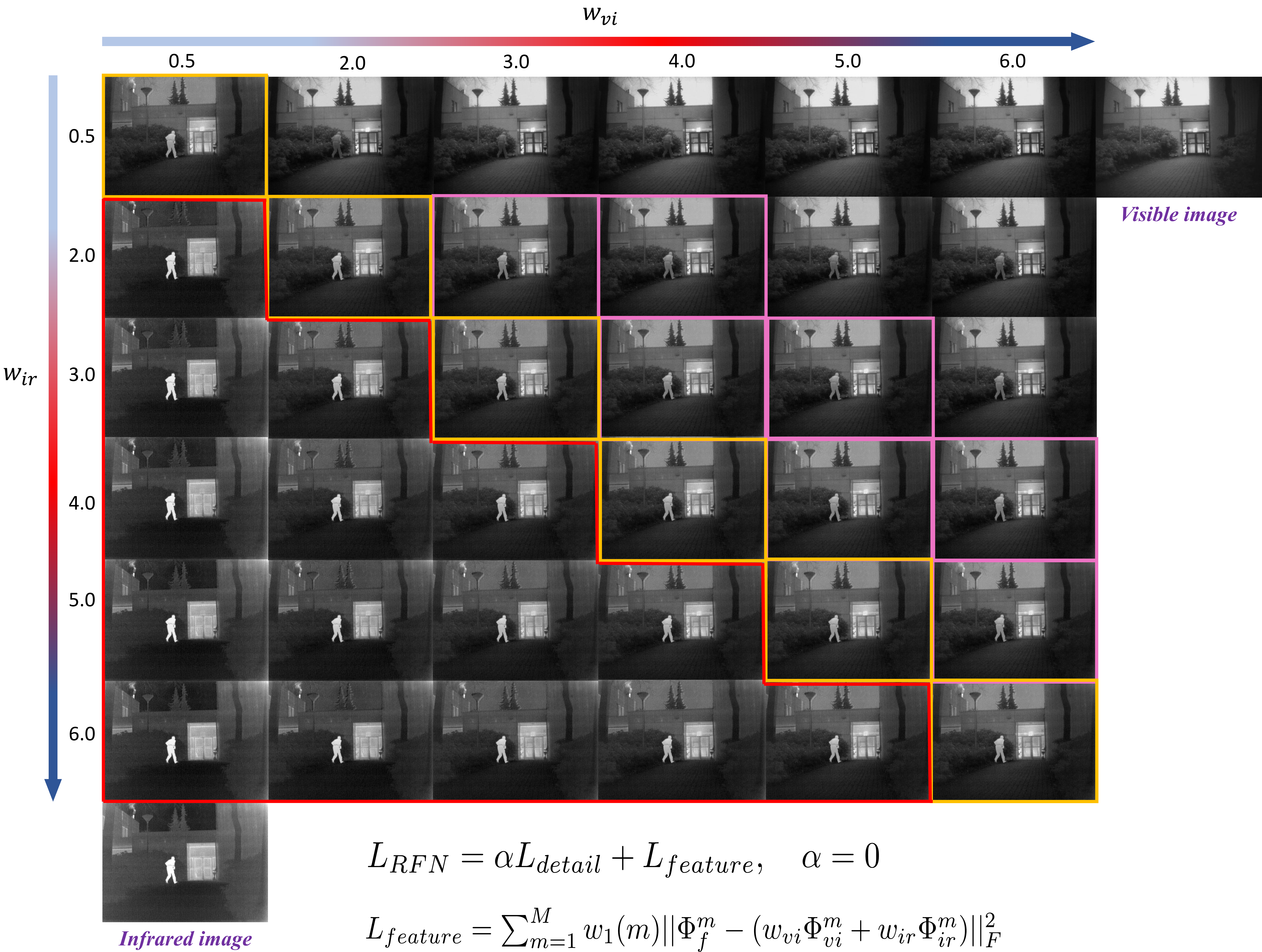}
	\caption{The fusion results obtained with different values of $w_{vi}$ and $w_{ir}$, when $\alpha=0$. In $L_{feature}$, $w_1\in{\{1,10,100,1000\}}$.}
	\label{fig:alpha0-wir-wvi}
\end{figure*}

\section{Experimental Validation}
\label{sec-experiments}
In this section, we conduct an experimental validation of the proposed fusion method. After detailing the experimental settings in the training phase and the test phase, we present several ablation studies to investigate the effect of different elements of the proposed fusion network. Finally, we compare our fusion framework with other existing algorithms qualitatively. For this purpose, we use several performance metrics to evaluate the fusion performance objectively.

Our network is implemented on the NVIDIA TITAN Xp GPU using PyTorch as a programming environment.

\begin{figure*}[!ht]
	\centering
	\includegraphics[width=\linewidth]{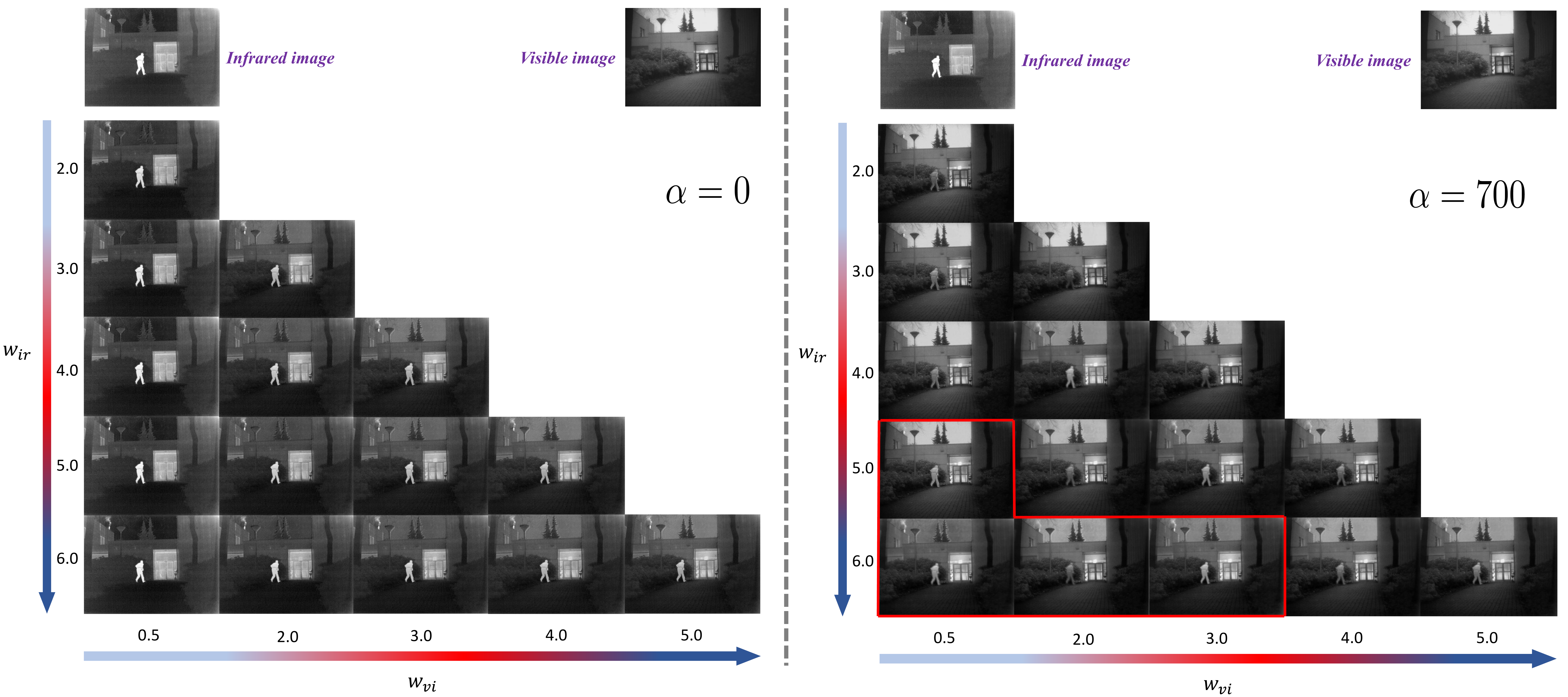}
	\caption{The fusion results obtained without ($\alpha=0$) or with ($\alpha=700$) $L_{detail}$. In $L_{feature}$, $w_1\in{\{1,10,100,1000\}}$.}
	\label{fig:alpha0-1e4-wir-wvi}
\end{figure*}

\subsection{Experimental Settings in the Training Phase}
\label{train-data}
In this section, we introduce the training datasets used in our two-stage training strategy. 

In the first stage,  we use the dataset MS-COCO \citep{lin2014microsoft} to train our auto-encoder network. 80000 images are chosen to constitute the training set. These images are converted to gray scale and reshaped to $256\times 256$. In Eq. \ref{equ:loss-auto}, the parameter $\lambda$ is set to 100 to balance the magnitude difference between $L_{pixel}$ and $L_{ssim}$. The batch size and epoch are set to 4 and 2, respectively. The learning rate is set to $1\times 10^{-4}$.

For the second training stage, we choose the KAIST \citep{hwang2015multispectral} dataset to train our RFN networks. It contains almost 90000 pairs of images. In this dataset, 80000 pairs of infrared and visible images are chosen for training. These images are also converted to gray scale and resized to $256\times 256$. The batch size and epoch are set to 4 and 2, respectively. The learning rate was also set to $1\times 10^{-4}$, as in the first stage.

\subsection{Experimental Settings in the Test Phase}
Our test images come from two datasets which were collected from TNO \citep{tno2014} and VOT2020-RGBT  \citep{vot2020rgbt}. These images are available at \citep{rfn2020}. Some samples of these images are shown in Fig.\ref{fig:example}. The first dataset contains 21 pairs of infrared and visible images collected from TNO. The second dataset contains 40 pairs of infrared and visible images, which were collected from TNO and VOT2020-RGBT. 


We use six quality metrics\footnote{The definitions of these metrics are introduced in our supplementary material.} to evaluate  our fusion algorithm objectively. These include: entropy($En$) \citep{roberts2008assessment}; standard deviation ($SD$) \citep{rao1997fibre}; mutual information($MI$) \citep{qu2002information}; modified fusion artifacts measure($N_{abf}$) \citep{kumar2013multifocus}, which evaluates the noise information in fused images; the sum of the correlations of differences($SCD$) \citep{aslantas2015new}; and the multi-scale structural similarity(MS-SSIM) \citep{ma2015perceptual}. The fusion performance improves with the increasing numerical index of all these seven metrics.


\subsection{Ablation Study for $L_{detail}$ and $L_{feature}$}
\label{abs-para-w}
In this section, we discuss the effect of $L_{detail}$ and tune the parameters in $L_{feature}$. Then, we investigate the impact of the relative weights of the visible and infrared features on the fusion performance.

Once the auto-encoder network is trained in the first stage, the parameters of the encoder and decoder are fixed and we use $L_{RFN}$ to train four RFN networks. As discussed in Section \ref{sec-train-rfn}, due to the magnitude difference between $L_{detail}$ and $L_{feature}$, the value of the parameter $\alpha$ should be large. Furthermore, the role of $L_{detail}$ is to preserve the detail information from the visible image. Based on the above considerations, in this experiment, $\alpha$ is set to 0 and 700 to analyze its influence on our network.

In $L_{feature}$, $w_1$ is a trade-off vector to balance the $L_{feature}$ values between the scales. To preserve the salient features from the infrared image, $w_{vi}$ and $w_{ir}$ should be set appropriately. In view of the role of $L_{detail}$, $w_{vi}$ should be relatively small to reduce any  redundancy in reconstructing the image detail information. In contrast, $w_{ir}$ should be large to preserve the complementary salient features in the infrared image. However, if $w_{vi}$ is set to 0, which constrains the fused features to mirror the infrared features, the network fails to converge due to the conflicting constraints of $L_{detail}$ and $L_{feature}$. So, in our experiment, $w_{vi}$ is set to a non-zero value.

As different combination of $w_{vi}$ and $w_{ir}$ can lead to different fusion results, we analyze the influence of these two parameters for different values from the range of [0.5, 6.0]. 

Firstly, when $\alpha=0$, which means only $L_{feature}$ is utilized to train RFN networks, some of the fusion results with different $w_{vi}$ and $w_{ir}$ are shown in Fig.\ref{fig:alpha0-wir-wvi}.

\begin{table*}[ht]
	\scriptsize
	\centering
	\caption{\label{tab:para-detail}The average values of the objective metrics obtained with different parameters ($\alpha$, $w_{vi}$, $w_{ir}$) on 21 pairs of infrared and visible images.}
	\resizebox{0.9\linewidth}{!}{
		\begin{tabular}{|c|c|c|c|c|c|c|c|c|}
			\hline
			$\bm{\alpha}$ & $w_{ir}$ &$w_{vi}$ & $En$\citep{roberts2008assessment} & $SD$\citep{rao1997fibre} & $MI$\citep{qu2002information} & $N_{abf}$\citep{kumar2013multifocus} & $SCD$\citep{aslantas2015new} & MS-SSIM\citep{ma2015perceptual} \\
			\hline
			\multirow{13}*{0} &
			0.5 & 0.5	&6.71845 	&67.66313 	&13.43690 	&0.09354 	&1.83520 	&0.92903 \\
			\cline{2-9}
			
			~& \multirow{3}*{2.0} &
			2.0 & 6.71557 	&67.63524 	&13.43114 	&0.09252 	&1.83495 	&0.92887 \\
			~& ~&3.0 & 6.80410 	&74.73724 	&13.60821 	&0.09240 	&1.82712 	&0.92294 \\
			~& ~&4.0 & 6.83492 	&\emph{\color{red}{79.75125}} 	&13.66983 	&0.09419 	&1.78649 	&0.90543 \\
			
			\cline{2-9}
			
			~& \multirow{3}*{3.0} &
			3.0 &  6.72263 	&67.83451 	&13.44526 	&0.09339 	&\textbf{1.83713} 	&\textbf{0.92988} \\
			~& ~&4.0 &  6.78738 	&72.45840 	&13.57476 	&0.09230 	&1.83518 	&0.92715 \\
			~& ~&5.0 &  6.81292 	&76.36078 	&13.62583 	&0.09324 	&1.81501 	&0.91696 \\
			
			\cline{2-9}
			
			~& \multirow{3}*{4.0} &
			4.0 &  6.72150 	&67.53190 	&13.44299 	&0.09355 	&1.83367 	&0.92842 \\
			~& ~&5.0 &  6.77188 	&70.98434 	&13.54376 	&0.09208 	&1.83538 	&0.92775 \\
			~& ~&6.0 &  6.80239 	&74.64694 	&13.60478 	&0.09218 	&1.82594 	&0.92263 \\
			
			\cline{2-9}
			
			~& \multirow{2}*{5.0} &
			5.0 &  6.71684 	&67.48675 	&13.43368 	&0.09218 	&1.83366 	&0.92847 \\
			~& ~&6.0 &  6.76875 	&70.35820 	&13.53750 	&0.08944 	&\emph{\color{red}{1.83707}} 	&0.92870 \\
			
			\cline{2-9}
			~& 	6.0 & 6.0	&6.72585 	&67.82480 	&13.45170 	&0.09209 	&1.83665 	&\emph{\color{red}{0.92949}} \\
			\Xhline{1pt}
			
			\multirow{3}*{700} &
			5.0 & 0.5	&\textbf{6.95916} 	&\textbf{91.41847} 	&\textbf{13.9183} 	&0.14375 	&1.58717 	&0.84109 \\
			\cline{2-9}
			~& \multirow{2}*{6.0} &
			0.5	&6.79112 	&68.28532 	&13.58224 	&\emph{\color{red}{0.07838}} 	&1.78391 	&0.88602 \\
			~& ~& 	 3.0	&\emph{\color{red}{6.84134}} 	&71.90131 	&\emph{\color{red}{13.68269}} 	&\textbf{0.07288} 	&1.83676 	&0.91456 \\
			\hline
	\end{tabular}}
\end{table*}

In Fig.\ref{fig:alpha0-wir-wvi}, when $w_{ir}$ is small, the fused images are similar to the visible image and the salient features in the infrared images are suppressed (as shown in first two rows). On the contrary, when $w_{ir}$ is large  (greater than 3.0), the salient features in the infrared image are retained. In contrast, the detail information in the visible image is not preserved.

To capture both types of information, for $\alpha=0$, we choose a middle value (yellow and pink boxes in Fig.\ref{fig:alpha0-wir-wvi}) to perform the objective evaluation. The  evaluation metrics  for different $w_{vi}$ and $w_{ir}$ are presented in Table \ref{tab:para-detail}. The best values are indicated in \textbf{bold}.

When $\alpha=700$, the detail information is preserved by $L_{detail}$. The aim of $L_{feature}$ is to promote  the salient features conveyed by the  source images. Accordingly, the values of $w_{vi}$ must be smaller than $w_{ir}$. We choose different combinations of $w_{vi}$ and $w_{ir}$ (the red boxes in Fig.\ref{fig:alpha0-wir-wvi}) to find the best values of $w_{vi}$ and $w_{ir}$. The fusion results obtained with ($\alpha=700$) or without ($\alpha=0$) $L_{detail}$ in the same combinations of $w_{vi}$ and $w_{ir}$ are shown in Fig.\ref{fig:alpha0-1e4-wir-wvi}.

In Fig.\ref{fig:alpha0-1e4-wir-wvi} (right part), the fusion results in the red boxes contain more detail information from the source images, yet the infrared features are still maintained. Compared with the left part, the fusion results on the right (red boxes) evidently preserve more detail information. When $\alpha=700$, the objective values for different parameters (Fig.\ref{fig:alpha0-1e4-wir-wvi} (right part), red boxes) are also presented in Table \ref{tab:para-detail}.

From Fig.\ref{fig:alpha0-wir-wvi} and Table \ref{tab:para-detail}, the different values of $\alpha$, $w_{vi}$ and $w_{ir}$ have a significant influence on the results. If the detail preservation loss function ($L_{detail}$) is not used ($\alpha=0$) in the training phase, the proposed fusion network fails to obtain acceptable fusion results. Although the fusion performance appears to be comparable in subjective evaluation (Fig.\ref{fig:alpha0-wir-wvi}, yellow and pink boxes), the subjective and objective assessments indicates  a notable degradation compared with the optimal parameter combination ($\alpha=700$, $w_{ir}=6.0$ and $w_{vi}=3.0$).

When the detail preservation loss function ($L_{detail}$) is switched on $\alpha=700$, our RFN-Nest fusion network scores the comparable metrics values of six metrics with $w_{ir}=6.0$ and $w_{vi}=3.0$. Based on this analysis, we set $w_{ir}=6.0$ and $w_{vi}=3.0$  in our next experiments. 

In next section, we will analyze the impact of parameter $\alpha$ in our loss function.

\subsection{Ablation Study for $\alpha$ in $L_{RFN}$}

As discussed in Section \ref{abs-para-w}, when the detail preserving loss function $L_{detail}$ is discarded  ($\alpha=0$), both the subjective and objectively measured   fusion performance will be poor. It is evident from from Fig.\ref{fig:alpha0-1e4-wir-wvi} and Table \ref{tab:para-detail} that our fusion network can achieve better fusion performance when $\alpha$ is not 0. Thus, choosing an optimal value of $\alpha$ becomes an important issue.

In our study, the parameters of $w_{vi}$ and $w_{ir}$ are set to 3.0 and 6.0, respectively. $w_1$ is set to $\{1, 10, 100, 1000\}$  to balance the discrepancy in the orders of magnitude of different scales. To find the optimal $\alpha$, we set it to $\{10, 100, 200, 500, 700, 1000\}$ and compute the results.

\begin{figure*}[!ht]
	\centering
	\includegraphics[width=0.9\linewidth]{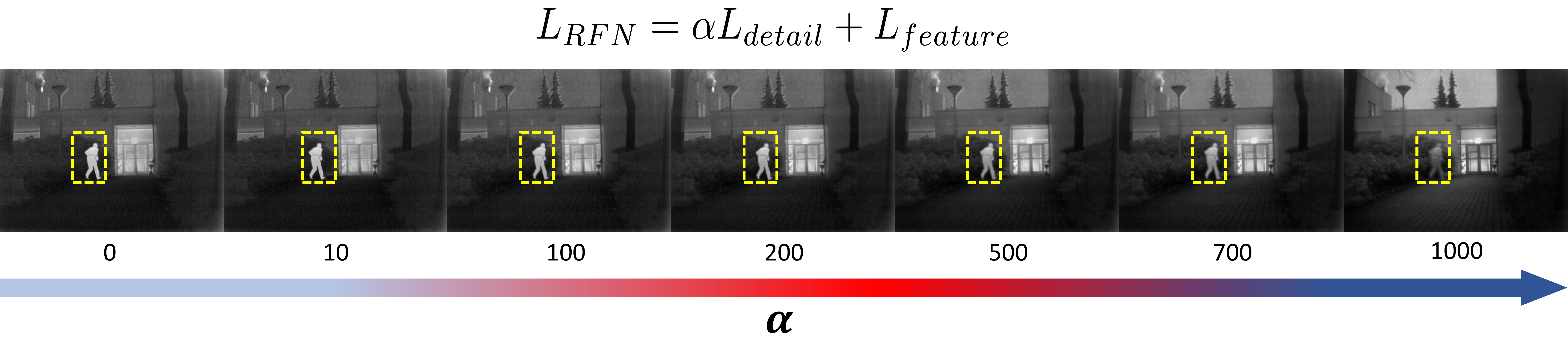}
	\caption{The fusion results obtained with different $\alpha$.}
	\label{fig:alpha}
\end{figure*}

Some examples of the fusion results are sown in Fig.\ref{fig:alpha}. With the increase  $\alpha$ (1000), the  salient features (man in the yellow box) are not clear, even suppressed, which makes the fused image similar to the visible image. When $\alpha$ is set to 500 and 700, the fusion results contain more detail information and the salient features are also maintained.

Based on these  observations, we objectively evaluate our fusion method with $\alpha$  set to 10, 100, 200, 500, 700, 1000. The metrics values of the fusion results with different $\alpha$ are shown in Table \ref{tab:abs-alpha}. The best values are indicated in \textbf{bold}.


\begin{table*}[!ht]
	\scriptsize
	\centering
	\caption{\label{tab:abs-alpha} The average metrics values of the proposed fusion network with different $\alpha$ on 21 pairs of infrared and visible images.}
	\resizebox{0.85\linewidth}{!}{
		\begin{tabular}{|c|c|c|c|c|c|c|}
			\hline
			$\bm{\alpha}$	& $En$\citep{roberts2008assessment} & $SD$\citep{rao1997fibre} & $MI$\citep{qu2002information} & $N_{abf}$\citep{kumar2013multifocus} & $SCD$\citep{aslantas2015new} & MS-SSIM\citep{ma2015perceptual}\\
			\hline
			10	&6.66878 	&62.83593 	&13.33757 	&0.08012 	&1.77192 	&0.87593 \\
			\hline
			100	&6.70939 	&63.77416 	&13.41878 	&0.07129 	&1.79329 	&0.90096 \\
			\hline
			200	&6.75446 	&66.01632 	&13.50893 	&\textbf{0.06680} 	&1.80880 	&0.91177 \\
			\hline
			500	&6.82103 	&70.34117 	&13.64206 	&0.06768 	&1.83252 	&0.91453 \\
			\hline
			700	&\textbf{6.84134} 	&\textbf{71.90131} 	&\textbf{13.68269} 	&0.07288 	&\textbf{1.83676} 	&\textbf{0.91456} \\
			\hline
	\end{tabular}}
\end{table*}

As shown in Fig.\ref{fig:alpha} and Table \ref{tab:abs-alpha}, when $\alpha$ is 700, the proposed fusion network achieves better fusion performance in both subjective and objective evaluation. In Table \ref{tab:abs-alpha}, the proposed network scores best in four out of seven metrics with $\alpha=700$. Thus, in our next experiments, the parameter $\alpha$ is set to 700.

\newpage
\subsection{Ablation Study for Training Strategy}

The proposed two-stage training strategy is a critical operation in our training phase. In this section, we discuss why this strategy is effective, and show its relative merits compared to the one-stage strategy.

\begin{figure}[ht]
	\centering
	\includegraphics[width=\linewidth]{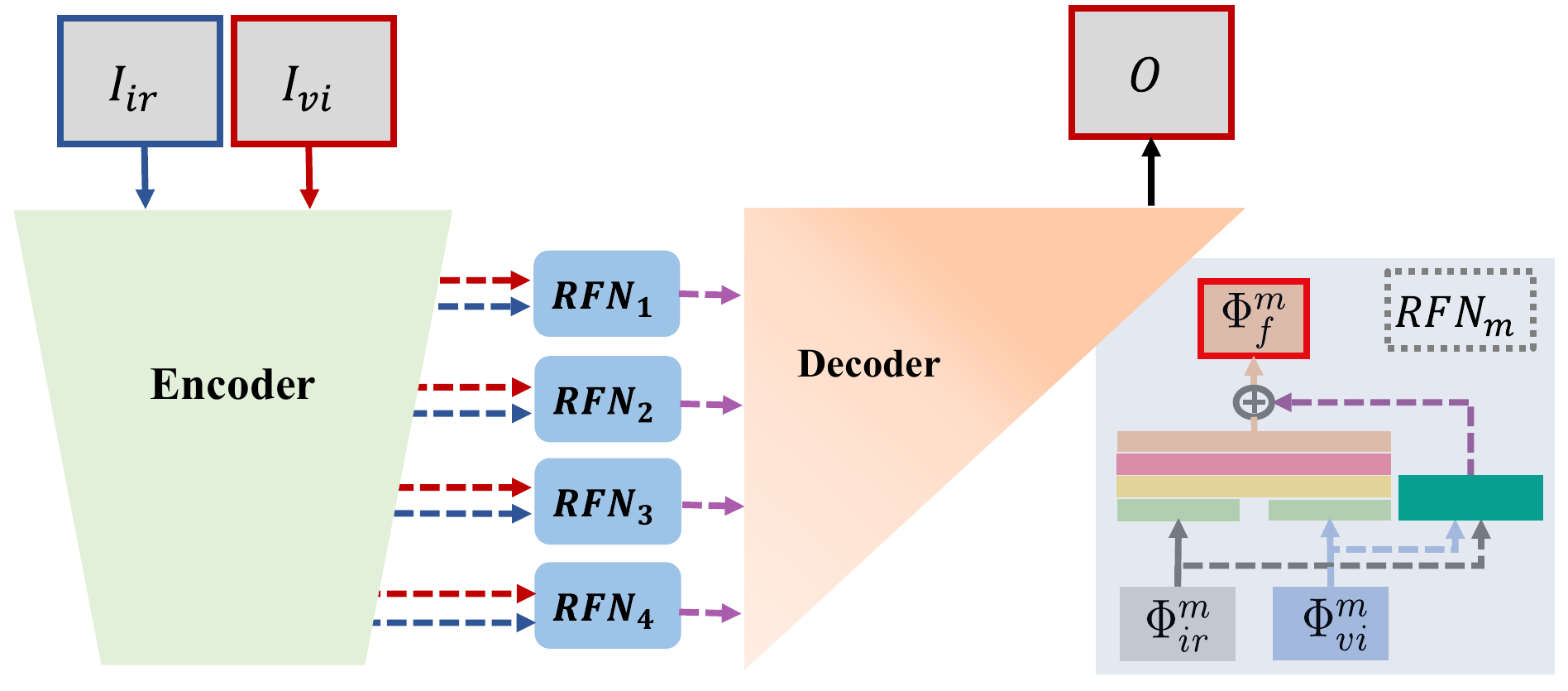}
	\caption{The training framework for the one-stage training strategy.}
	\label{fig:one-stage}
\end{figure}

\begin{figure}[ht]
	\centering
	\includegraphics[width=\linewidth]{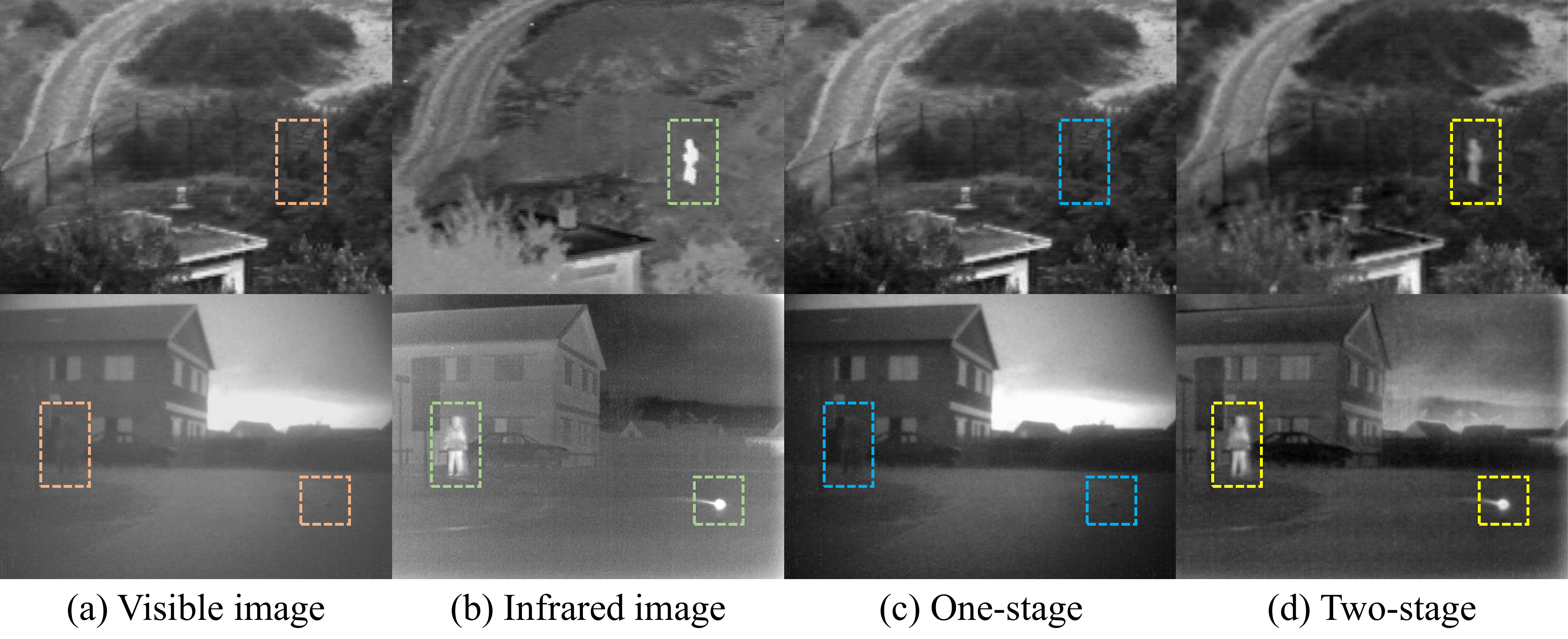}
	\caption{The fusion results obtained by one-stage and two-stage training strategy. (a) Visible images; (b) Infrared images; (c) Fused images obtained by one-stage strategy; (d) Fused images obtained by two-stage strategy.}
	\label{fig:onestage-nest-results}
\end{figure}

One-stage training strategy means the encoder, RFN and decoder are trained, simultaneously. The training framework is shown in Fig.\ref{fig:one-stage}, where both the encoder and the decoder are free to adapt their weights. The loss function and the parameter settings are the same as $L_{RFN}$, which means $\alpha=700$, $w_1=\{1,10,100,1000\}$, $w_{vi}=3.0$ and $w_{ir}=6.0$. The fusion results obtained by these two training strategies are shown in Fig.\ref{fig:onestage-nest-results}.

In Fig.\ref{fig:onestage-nest-results} (c), the visible spectrum detail information is enhanced with the one-stage training strategy. However, the salient objects in infrared image are lost. The premise of image fusion is not realised. In contrast, the two-stage training strategy (Fig.\ref{fig:onestage-nest-results}, d) enables  the fused image to preserve the salient infrared objects and contain more detail information from visible images.

The reason is that the encoder and the decoder may not have the desirable feature extraction and reconstruction ability when designed using the one-stage training strategy. More importantly, as the RFN is the key in our fusion network, it should be trained carefully to obtain good fusion performance. 

In conclusion, we use the two-stage training strategy to train our fusion network. In the first training stage, the encoder is trained to extract powerful multi-scale deep features, to be used by the decoder for image reconstruction. In the second stage, with the fixed encoder and decoder, the RFN networks are trained to fuse the multi-scale deep features, to enhance the detail information from the visible spectrum image  and to preserve salient features from the infrared source image.

\subsection{Ablation Study for Nest Connection in Decoder}

\begin{figure}[ht]
	\centering
	\includegraphics[width=0.75\linewidth]{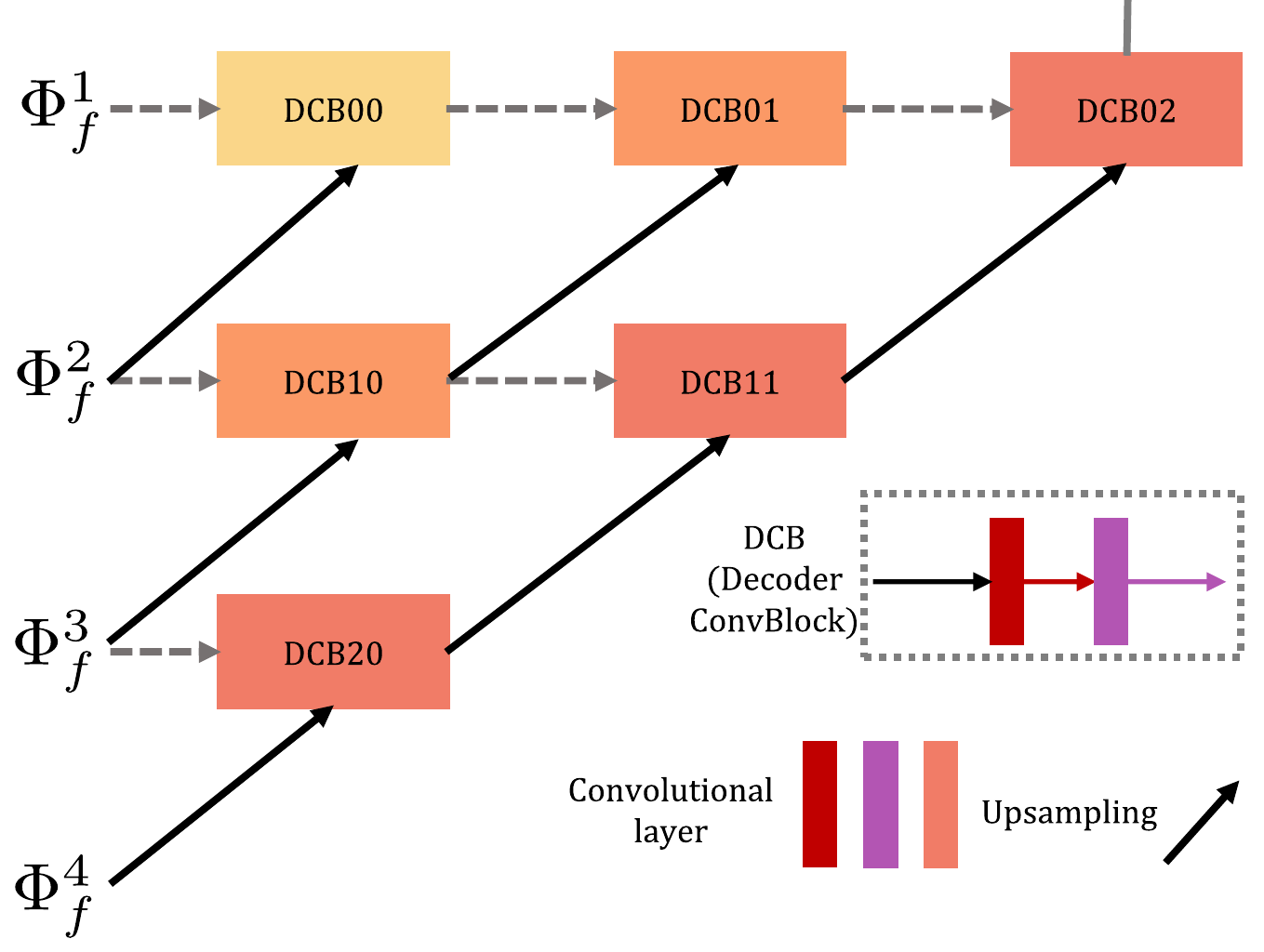}
	\caption{The decoder network without short connection (``\emph{No-nest}'').}
	\label{fig:no-short-connection}
\end{figure}

In this section, we discuss the influence of the nest connection in the decoder. Fig.\ref{fig:no-short-connection} shows the decoder network structure without nest connection (remove the short connections between ``DCB''). We train this new decoder architecture with the same training strategy and the same loss functions as discussed in Section \ref{strategy}. 

\begin{table*}[ht]
	\scriptsize
	\centering
	\caption{\label{tab:abs-fuison-strategy}The quality metrics values of two ablation studies. ``\emph{No-nest}'' indicates the decoder without nest-connection architecture. ``Encoder \& Decoder'' denotes that the encoder and the decoder are fixed, the fusion strategy is ``$add$'', ``$max$'', ``$l_1$-norm'', ``$l_*$-norm'' or ``SCA''. ``RFN-Nest'' means the proposed fusion network.}
	\resizebox{0.95\linewidth}{!}{
		\begin{tabular}{|c|c|c|c|c|c|c|c|}
			\hline
			\multicolumn{2}{|c|}{} & $En$\citep{roberts2008assessment} & $SD$\citep{rao1997fibre} & $MI$\citep{qu2002information} & $N_{abf}$\citep{kumar2013multifocus} & $SCD$\citep{aslantas2015new} & MS-SSIM\citep{ma2015perceptual} \\
			\hline
			
			\multicolumn{2}{|c|}{\emph{No-nest}}		
			&6.75935 	&66.48558 	&13.51871 	&\textbf{0.05278} 	&1.80356 	&0.90172 \\
			\hline
			\multirow{5}*{\makecell[c]{Encoder \\ \& \\ Decoder}} &	
			$add$		&6.68274 	&67.45593 	&13.36548 	&0.09209 	&\emph{\color{red}{1.83367}} 	&\textbf{0.92831} \\
			\cline{2-8}
			&$max$		&6.71760 	&\emph{\color{red}{92.49952}} 	&13.43519 	&0.21454 	&1.58628 	&0.77823 \\
			\cline{2-8}
			&$l_1$-norm	&\emph{\color{blue}{6.83073}} 	&\textbf{93.21573} 	&\emph{\color{blue}{13.66146}} 	&0.20760 	&1.56378 	&0.76769 \\
			\cline{2-8}
			&$l_*$-norm	&6.81192 	&73.66134 	&13.62385 	&\emph{\color{blue}{0.09010}} 	&\emph{\color{blue}{1.80934}} 	&\emph{\color{red}{0.92628}} \\
			\cline{2-8}
			&SCA		&\textbf{6.91971} 	&\emph{\color{blue}{82.75242}} 	&\textbf{13.83942} 	&0.13405 	&1.73353 	&0.86248 \\
			\hline
			\multicolumn{2}{|c|}{RFN-Nest}		&\emph{\color{red}{6.84134}} 	&71.90131 	&\emph{\color{red}{13.68269}} 	&\emph{\color{red}{0.07288}} 	&\textbf{1.83676} 	&\emph{\color{blue}{0.91456}} \\
			\hline 
	\end{tabular}}
\end{table*}


The values of seven quality metrics are shown in Table \ref{tab:abs-fuison-strategy}. The ``\emph{No-nest}'' denotes the decoder without nest-connection architecture. The best values, the second-best values and the third-best values are indicated in \textbf{bold}, \emph{\color{red}{red and italic}} and \emph{\color{blue}{blue and italic}}, respectively.

Compared with ``\emph{No-nest}'', the RFN-Nest (the decoder with nest connection) obtains one best metrics value, three second-best metrics values and one third-best metrics value. This indicates that the nest connection architecture plays an important role in  boosting the reconstruction ability of the decoder network. With the nest connection, the decoder is able to preserve  more image information conveyed by the multiscale deep features ($MI$, $FFMI_{dct}$, $FFMI{w}$) and generate more natural and clearer fused image ($EN$, $SD$, $VIF$).


\subsection{Ablation Study for Fusion Strategy}
\label{sec-abla-rfn}

In this section, we analyze the importance of RFN as an adaptive fusion mechanism in our fusion network. We choose five classical handcrafted fusion strategies (``$add$'', ``$max$'', ``$l_1$-norm'', ``$l_*$-norm'' and ``SCA'') which are used in existing fusion networks \citep{li2018densefuse}\citep{zhang2020ifcnn}\citep{li2020nestfuse} to do the experiments.

The trained encoder and decoder are utilized to extract the multi-scale deep features and generate the final image from the fused features, respectively.

Let $\Phi_{ir}^m$ and $\Phi_{vi}^m$ denote the multi-scale deep features extracted by the trained encoder from the infrared and visible image, respectively. $\Phi_{f}^m$ are the fused deep features. $m$ indicates the scale of the deep features. The formulas of these five strategies are shown in Table \ref{tab:fusion-strategy}.

\begin{table}[ht]
\footnotesize
\centering
\caption{\label{tab:fusion-strategy} The formulas of different fusion strategies.}
\resizebox{0.85\linewidth}{!}{
\begin{tabular}{|c|l|}
\hline
Fusion Strategy & Formula \\
\hline
$add$\citep{li2018densefuse} & $\Phi_{f}^m = \Phi_{ir}^m + \Phi_{vi}^m$ \\
\hline
$max$\citep{zhang2020ifcnn} & $\Phi_{f}^m = \max(\Phi_{ir}^m, \Phi_{vi}^m)$ \\
\hline 
$l_1$-norm\citep{li2018densefuse} & $\Phi_{f}^m = l_1(\Phi_{ir}^m, \Phi_{vi}^m)$ \\
\hline 
$l_*$-norm\citep{li2020nestfuse} & $\Phi_{f}^m = l_*(\Phi_{ir}^m, \Phi_{vi}^m)$ \\
\hline 
SCA\citep{li2020nestfuse} & \makecell[c]{Spatial and channel \\ attention fusion strategy} \\
\hline 
\end{tabular}}
\end{table}

 ``$add$'', means the fused features are obtained by adding the source features, directly. In ``$max$'' strategy, $max(\cdot)$ denotes an element wise choose-max strategy \citep{zhang2020ifcnn}. For the ``$l_1$-norm'' strategy, $l_1(\cdot)$, the weights are calculated based on $l_1$-norm. For details on how to calculate these weights, please refer to \citep{li2018densefuse}. For ``$l_*$-norm'' (known as nuclear-norm), $l_*(\cdot)$ calculates the sum of singular values of a matrix involved in the global pooling operation of deep features to obtain the fusion weights. 

\begin{figure*}[ht]
	\centering
	\includegraphics[width=0.9\linewidth]{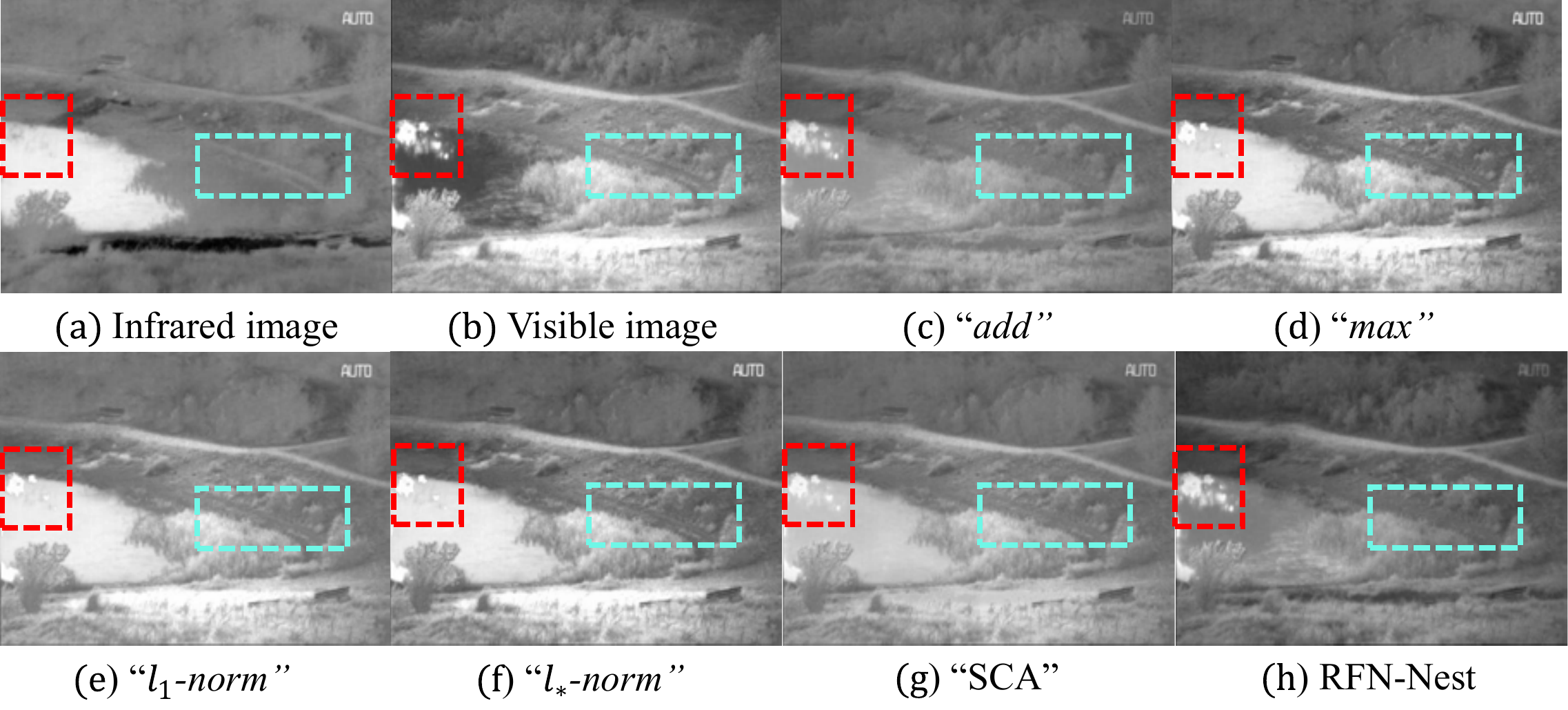}
	\caption{The fusion results with different fusion strategies. ``RFN-Nest'' means the adaptive RFN networks are utilized in the fusion operation.}
	\label{fig:fusion-strategy}
\end{figure*}

The ``SCA'' indicates the Spatial/Channel Attention fusion strategy which was utilized in NestFuse \citep{li2020nestfuse}. In this experiment\footnote{The ``SCA'' fusion strategy is used in NestFuse.}, the $l_1$-norm is used to do the spatial attention fusion and the average pooling is utilized to calculate the channel attention.

Some examples of the images fused using different fusion strategies are shown in Fig.\ref{fig:fusion-strategy}. Compared with other handcrafted fusion strategies, the fused image obtained by the RFN-based network preserves more detail information from visible image (blue boxes) and the fused image contains less artefacts (red boxes).

The results of fusing 21 pairs of infrared and visible images have been evaluated in terms of the seven quality metrics. The metrics values are shown in Table \ref{tab:abs-fuison-strategy}. The table also reports the results obtained  with other fusion strategies. The RFN-based network (RFN-Nest) achieves five best values. This indicates that when the learnable fusion network is used as a fusion strategy, the detail image information will be boosted ($En$, $SD$) thanks to the proposed loss function. Regarding feature preservation, the proposed strategy still obtains three best values ($MI$, $FFMI_{dct}$ and $FFMI_{w}$) and two comparable results ($SSIM_a$ and $VIF$).

In Section \ref{rgbt-tracking}, we adopt this learnable fusion network (RFN) for the  object tracking task to illustrate the effectiveness of RFN-based fusion strategy in other vision task.


\subsection{Fusion Results Analysis on 21 pairs Images}
\label{fusion-result}

To compare the fusion performance of the proposed method with the state-of-the-art algorithms, eleven representative fusion methods are chosen, including discrete cosine harmonic wavelet transform(DCHWT) \citep{kumar2013multifocus}, gradient transfer and total variation minimization(GTF) \citep{ma2016infrared}, convolutional sparse representation(ConvSR) \citep{liu2016image}, multi-layer deep features fusion method(VggML) \citep{li2018infrared}, DenseFuse \citep{li2018densefuse}, FusionGAN \citep{ma2019fusiongan}, IFCNN \citep{zhang2020ifcnn} (elementwise-maximum), NestFuse \citep{li2020nestfuse}, PMGI \citep{zhang2020PMGI}, DDcGAN \citep{ma2020ddcgan} and U2Fusion \citep{xu2020u2fusion}.

\begin{figure*}[!ht]
	\centering
	\includegraphics[width=\linewidth]{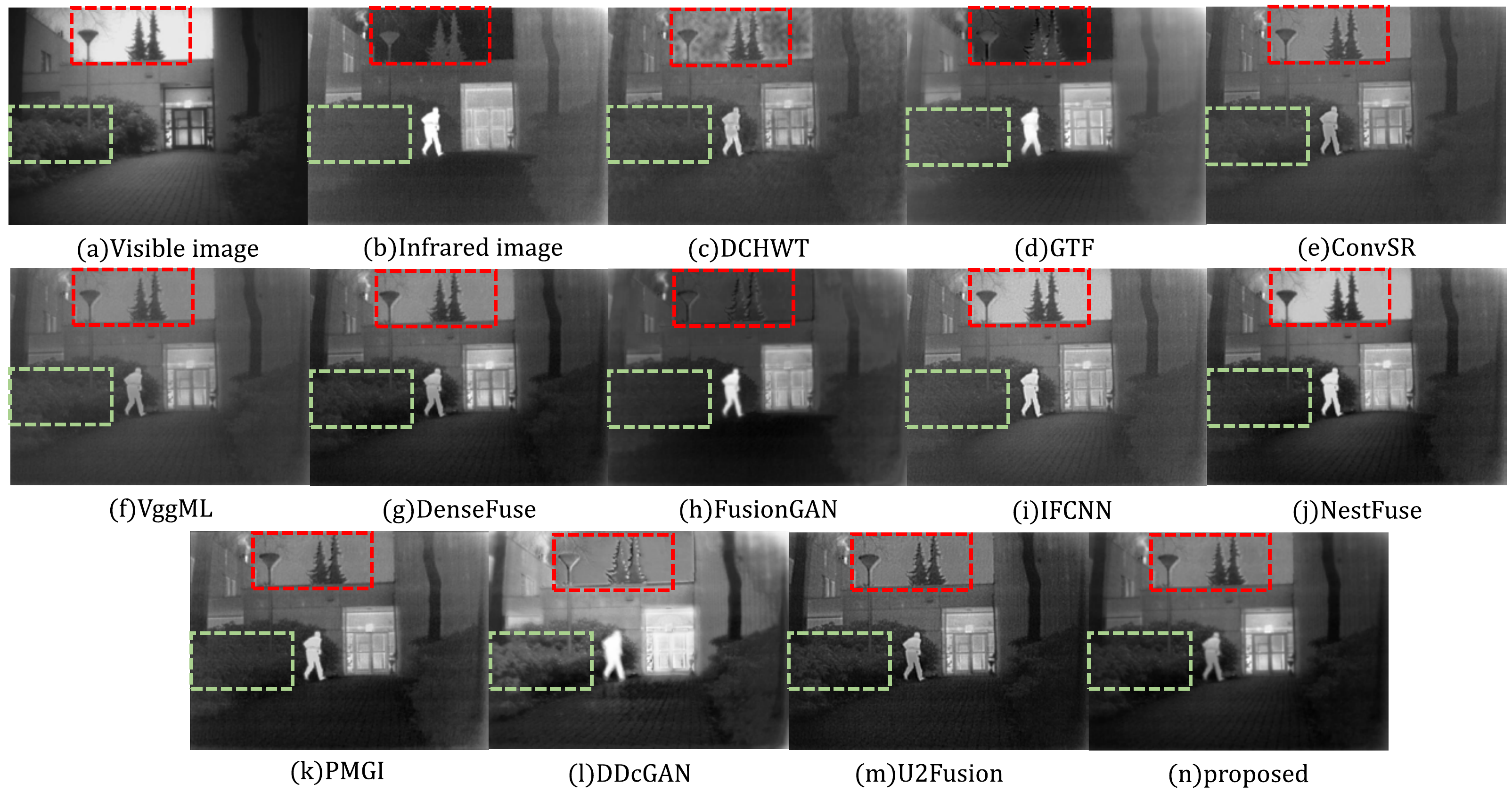}
	\caption{The experimental results on ``man'' images. (a) Visible; (b) Infrared; (c) DCHWT; (d) GTF; (e) ConvSR; (f) VggML; (g) DenseFuse; (h) FusionGAN; (i) IFCNN; (j) NestFuse; (k) PMGI; (l) DDcGAN; (m) U2Fusion; (n) proposed.}
	\label{fig:man}
\end{figure*}


\begin{figure*}[!ht]
\centering
\includegraphics[width=\linewidth]{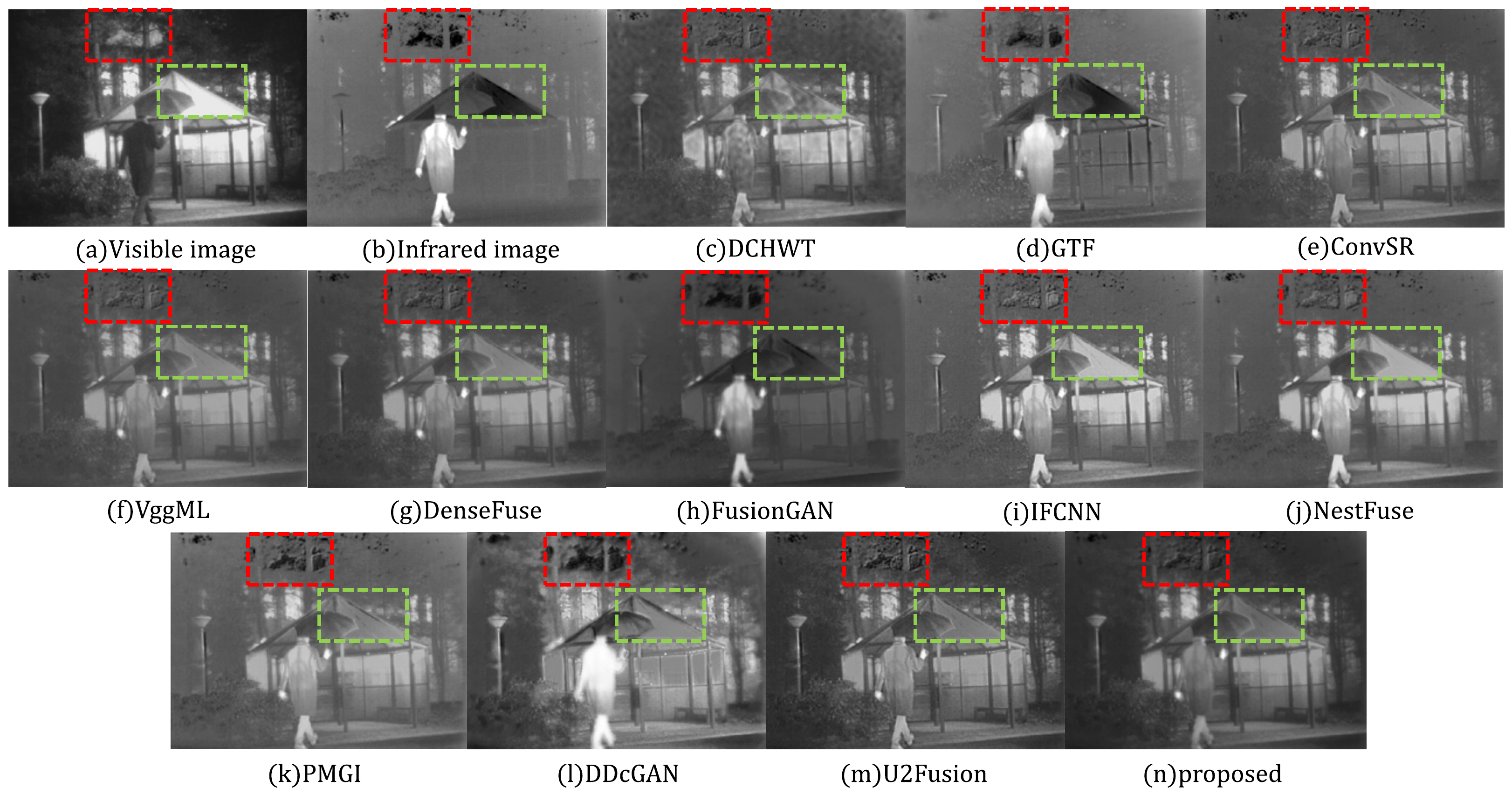}
\caption{The experimental results on ``umbrella'' images. (a) Visible; (b) Infrared; (c) DCHWT; (d) GTF; (e) ConvSR; (f) VggML; (g) DenseFuse; (h) FusionGAN; (i) IFCNN; (j) NestFuse; (k) PMGI; (l) DDcGAN; (m) U2Fusion; (n) proposed.}
\label{fig:umbrella}
\end{figure*}

\begin{table*}[!ht]
\scriptsize
\centering
\caption{\label{tab:results}The values of seven quality metrics averaged over the fused images on 21 pairs of infrared and visible images.}
\resizebox{0.9\linewidth}{!}{
\begin{tabular}{|c|c|c|c|c|c|c|}
	\hline
	~ & $En$\citep{roberts2008assessment} & $SD$\citep{rao1997fibre} & $MI$\citep{qu2002information} & $N_{abf}$\citep{kumar2013multifocus} & $SCD$\citep{aslantas2015new} & MS-SSIM\citep{ma2015perceptual} \\
	\hline
	DCHWT\citep{kumar2013multifocus}		&6.56777 	&64.97891 	&13.13553 	&0.12295 	&1.60993 	&0.84326 \\
	\hline
	GTF\citep{ma2016infrared}			&6.63433 	&67.54361 	&13.26865 	&0.07951 	&1.00488 	&0.80844 \\
	\hline
	ConvSR\citep{liu2016image}		&6.25869 	&50.74372 	&12.51737 	&0.01958 	&1.64823 	&0.90281 \\
	\hline
	VggML\citep{li2018infrared}		&6.18260 	&48.15779 	&12.36521 	&\textbf{0.00120} 	&1.63522 	&0.87478 \\
	\Xhline{1pt} 							
	DenseFuse\citep{li2018densefuse}	&6.67158 	&67.57282 	&13.34317 	&0.09214 	&\emph{\color{red}{1.83502}} 	&\textbf{0.92896} \\
	\hline
	FusionGan\citep{ma2019fusiongan}	&6.36285 	&54.35752 	&12.72570 	&\emph{\color{red}{0.06706}} 	&1.45685 	&0.73182 \\
	\hline
	IFCNN\citep{zhang2020ifcnn}		&6.59545 	&66.87578 	&13.19090 	&0.17959 	&1.71375 	&0.90527 \\
	\hline
	NestFuse\citep{li2020nestfuse}	&\emph{\color{blue}{6.91971}} 	&\emph{\color{red}{82.75242}} 	&\emph{\color{blue}{13.83942}} 	&0.13405 	&1.73353 	&0.86248 \\
	\hline
	PMGI\citep{zhang2020PMGI}		&\emph{\color{red}{6.93391}} 	&71.54806 	&\emph{\color{red}{13.86783}} 	&0.13525 	&1.78242 	&0.88934 \\
	\hline
	DDcGAN\citep{ma2020ddcgan}		&\textbf{7.47310} 	&\textbf{100.34809}  &\textbf{14.94620} 	&0.33784 	&1.60926 	&0.76636 \\
	\hline
	U2Fusion\citep{xu2020u2fusion}	&6.75708 	&64.91158 	&13.51416 	&0.29088 	&\emph{\color{blue}{1.79837}} 	&\emph{\color{red}{0.92533}} \\
	\hline
	\emph{\textbf{proposed}}		&6.84134 	&\emph{\color{blue}{71.90131}} 	&13.68269 	&\emph{\color{blue}{0.07288}} 	&\textbf{1.83676} 	&\emph{\color{blue}{0.91456}} \\
	\hline 
\end{tabular}}
\end{table*}

\begin{figure*}[!ht]
	\centering
	\includegraphics[width=0.9\linewidth]{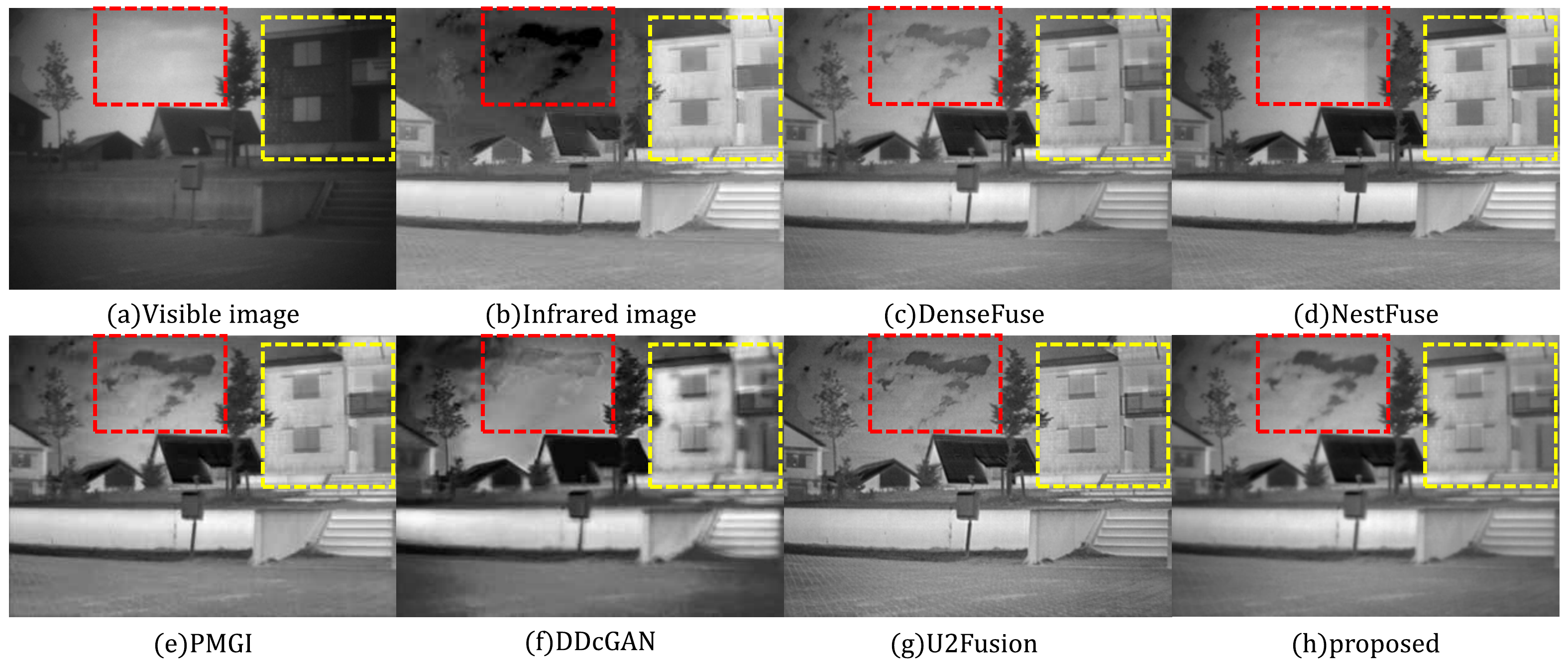}
	\caption{The experimental results on ``street'' images. (a) Visible; (b) Infrared; (c) DenseFuse; (d) NestFuse; (e) PMGI; (f) DDcGAN; (g) U2Fusion; (h) proposed.}
	\label{fig:street-40}
\end{figure*}

For DenseFuse, we choose the sum strategy and set the trade-off parameter to $1\times 10^2$. For NestFuse, the average pooling is utilized for the channel attention fusion strategy. All these fusion methods are implemented using publicly available codes, and their parameters are set by referring to their original reports.

To evaluate the visual effects of the fusion results\footnote{More experimental results are shown in our supplementary material.}, two pairs of visible and infrared images are selected, namely  ``man'' and ``umbrella''. The fused images obtained by the existing methods and our fusion method (RFN-Nest) are shown in Fig.\ref{fig:man} and Fig.\ref{fig:umbrella}, respectively.

In Fig.\ref{fig:man} and Fig.\ref{fig:umbrella}, the fused images obtained by DCHWT are more noisy and contain image artefacts. The fused images obtained by GTF and FusionGAN exhibit clearer features and more detailed background information. Although these fused images retain more complementary information, they look more like the infrared image, especially in the background. In view of the importance of the background information, ConvSR, VggML, DenseFuse, IFCNN, NestFuse, PMGI and U2Fusion are designed to preserve more detail information from the visible image. These methods appear to reduce the salient infrared features, compared with GTF and FusionGAN, producing acceptable fusion results. Although DDcGAN is also designed to maintain more detail information from visible images, in Fig.\ref{fig:man} (l), it injects more noise into the fused image and the infrared targets are blurred.

Although the target features are  not enhanced too much in the fused image, the contrast is better than in the visible image.  Moreover, for the detail preservation, in Fig.\ref{fig:man}, compared with the other fusion methods, the `tree' and `street lamp' (red box) are clearer in the fused image obtained by our proposed method. The detail textures of `bushes' (green box) are also preserved into the fused image. 

In Fig.\ref{fig:umbrella}, in the green box, many fusion methods are unable to preserve the salient features of `pavilion' from the visible image except IFCNN, NestFuse and the proposed method, which means these fusion methods fuse too much background information from the infrared image. Compared with all these fusion methods, in the red box, the detail information of the `tree' reconstructed by the proposed method is clearer in the fused image (Fig.\ref{fig:umbrella} (n)).

The background and the context around salient parts are not very clear and sometime even invisible because of the difficulty in extracting salient features from source images, as shown in Fig.\ref{fig:man} and Fig.\ref{fig:umbrella}. This drawback will cause a performance degradation when the image fusion algorithms are used in other computer vision tasks, such as RGB-T visual object tracking. In contrast, our RFN-Nest fusion network is able to preserve more detail information and to maintain the contrast of infrared parts.

Compared with all the above fusion methods, the fused image obtained by the proposed method appears to retain a better balance between the visible background information and the infrared features. 

We evaluate the fusion performance objectively using the seven quality metrics to compare the seventeen existing fusion methods and our proposed fusion framework. The values of these metrics averaged over all fused images are shown in Table \ref{tab:results}. The best values, the second-best values and the third-best values are indicated in \textbf{bold}, \emph{\color{red}{red and italic}} and \emph{\color{blue}{blue and italic}}, respectively.

From Table \ref{tab:results}, the proposed fusion framework (RFN-Nest) obtains one best values ($SCD$) and three third-best values ($SD$, $N_{abf}$, MS-SSIM) compared to the other methods. The reason why DDcGAN obtains larger values of $En$, $SD$ and $MI$ is that DDcGAN introduces more noise and artefacts into the fused image. Our fusion network achieves good fusion performance, producing sharper content and exhibiting more visual information fidelity. 


\begin{table*}[!ht]
	\scriptsize
	\centering
	\caption{\label{tab:results-40}The values of seven quality metrics averaged over the fused images on 40 pairs of infrared and visible images which collected from TNO and VOT2020.}
	\resizebox{0.95\linewidth}{!}{
		\begin{tabular}{|c|c|c|c|c|c|c|}
			\hline
			~& $En$\citep{roberts2008assessment} & $SD$\citep{rao1997fibre} & $MI$\citep{qu2002information} & $N_{abf}$\citep{kumar2013multifocus} & $SCD$\citep{aslantas2015new} & MS-SSIM\citep{ma2015perceptual} \\
			\hline 
			DenseFuse\citep{li2018densefuse}	&6.77630 	&73.63462 	&13.55261 	&\textbf{0.06346} 	&\emph{\color{red}{1.74862}} 	&\emph{\color{red}{0.92944}} \\
			\hline
			NestFuse\citep{li2020nestfuse}		&\emph{\color{red}{6.99347}} 	&\emph{\color{red}{90.28951}} 	&\emph{\color{red}{13.98693}} 	&\emph{\color{blue}{0.11138}} 	&1.67540 	&0.88611 \\
			\hline
			PMGI\citep{zhang2020PMGI}			&\emph{\color{blue}{6.96974}} 	&77.25462 	&\emph{\color{blue}{13.93948}} 	&0.11434 	&1.68523 	&0.88830 \\
			\hline
			DDcGAN\citep{ma2020ddcgan}			&\textbf{7.50173} 	&\textbf{106.99113}  &\textbf{15.00346} 	&0.30998 	&1.55359 	&0.78419 \\
			\hline
			U2Fusion\citep{xu2020u2fusion}		&6.94970 	&76.80347 	&13.89939 	&0.28363 	&\emph{\color{blue}{1.74780}} 	&\textbf{0.93141} \\
			\hline
			\emph{\textbf{proposed}}			&6.92952 	&\emph{\color{blue}{78.22247}} 	&13.85904 	&\emph{\color{red}{0.06357}} 	&\textbf{1.76116} 	&\emph{\color{blue}{0.90894}} \\
			\hline
	\end{tabular}}
\end{table*}

\subsection{Further Analysis on 40 Pairs Images}
\label{fusion-result-40}

The previous ablation studies and experiments are conducted on one test dataset which  contains 21 pairs of infrared and visible images. To verify the generalization performance of the proposed fusion network, a new test dataset is created. It contains 40 pairs of infrared and visible images which are collected from TNO \citep{tno2014} and VOT2020-RGBT \citep{vot2020rgbt}.

In this section, we choose several state-of-the-art deep learning based fusion methods to perform comparative experiments. These methods include DenseFuse \citep{li2018densefuse} which is a classical autoencoder-based fusion method, NestFuse \citep{li2020nestfuse} which has the same backbone (encoder and decoder) with the proposed method, and three latest fusion methods (PMGI \citep{zhang2020PMGI}, DDcGAN \citep{ma2020ddcgan} and U2Fusion \citep{xu2020u2fusion}).

An example of the fused images obtained by these fusion methods and the proposed network is shown in Fig.\ref{fig:street-40}. The unnatural textures in the sky are introduced by the infrared image (Fig.\ref{fig:street-40} (b), red box) should be looked more natural in the fused image. It is observed that DenseFuse, PMGI and the proposed method generate fused images of natural appearance. Moreover, compared with the existing fusion methods, our network also preserves more detail information from both infrared and visible images (Fig.\ref{fig:street-40}, the house in yellow box). Note, the DDcGAN again introduces noise into the fused image and blurs the salient feature content.

The same six quality metrics are used for comparative evaluation. The average values of these metrics are shown in Table \ref{tab:results-40}. The best values are indicated in \textbf{bold}, the second-best values are denoted in \emph{\color{red}{red and italic}} and the third-best values are denoted in \emph{\color{blue}{blue and italic}}. 

Compared with the results  on the 21 pairs of images, the proposed network exhibits even better performance on the 40 image pairs. The method achieves one best value ($SCD$), one second-best values ($N_{abf}$) and two third-best values ($SD$, MS-SSIM). Even compared with DDcGAN, our fusion network performance is comparable. This confirms that our fusion network trained by the two-stage fusion strategy and the novel loss function demonstrates better generalization.


\section{Experiments on RGBT Object Tracking}
\label{rgbt-tracking}

Over the past two years, multi-modality object tracking has been of interest  in many vision applications. In Vision Object Tracking challenge (VOT) 2019 \citep{vot2019rgbt}, for the first time, the committee introduced two new sub-challenges (RGBD and RGBT), in which each sequence in the dataset of RGBD or RGBT contains two modalities (RGB image and depth image, RGB image and infrared image) as the input. As we focus on the fusion of infrared and visible images, the RGBT sub-challenge data is used to evaluate the performance of the proposed learnable fusion network (RFN) and the novel loss functions.

In VOT2020 \citep{vot2020rgbt}, the video sequences are the same as VOT2019 \citep{vot2019rgbt}, but a new performance evaluation protocol is introduced for short-term tracker evaluation (includes RGBT sub-challenge). The new protocol avoids tracker-dependent resets and reduces the variance of the performance evaluation measures.

A state-of-the-art siamese-based tracker AFAT \citep{xu2020afat} is chosen to be the base tracker. In AFAT, a failure-aware system, realized by a Quality Prediction Network (QPN), based on convolutional and LSTM modules was proposed and obtained better tracking performance in many datasets. For RGBT object tracking, the proposed fusion strategy network (RFN) and the proposed loss function are incorporated  into AFAT.

\subsection{The RFN and The Loss Function For RGBT Tracking}

As we discussed, the proposed residual fusion network (RFN) is a learnable fusion strategy. Thus, ideally, when RFN is applied into AFAT \citep{xu2020afat}, it needs a sufficient quantity of data to train the whole model. 

\begin{figure}[!ht]
\centering
\includegraphics[width=0.85\linewidth]{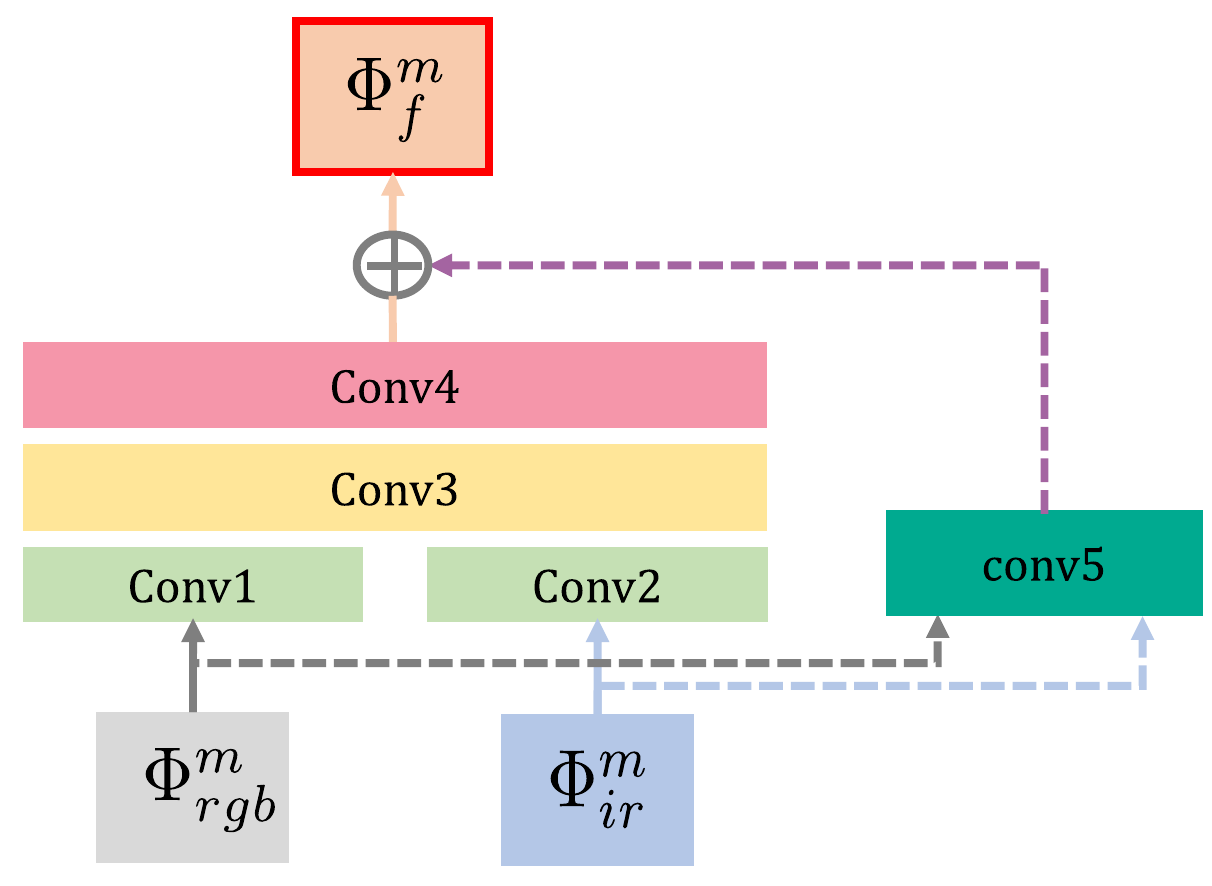}
\caption{The RFN architecture for RGBT tracking.}
\label{fig:rfn-tracking}
\end{figure}

However, due to the lack of  labeled training data, we were  forced to simplify the architecture of RFN, by reducing the number of  convolutional layers, as shown in Fig.\ref{fig:rfn-tracking}. In the training phase, we only train the RFN module, the AFAT modules are fixed to reduce the number of learnable parameters \footnote{The framework of RFN-base AFAT is shown in our supplementary material.}.

Three RGBT datasets are used to train our RFN module, namely GTOT \citep{li2016gtot}, VT821 \citep{tang2019vt821}, VT1000 \citep{tu2019vt1000}. These datasets only contain 17.6k frames in total. GTOT is a dataset for RGBT tracking, the VT821 and the VT1000 are built for RGBT salient object detection.

\begin{figure*}[!ht]
	\centering
	\includegraphics[width=0.9\linewidth]{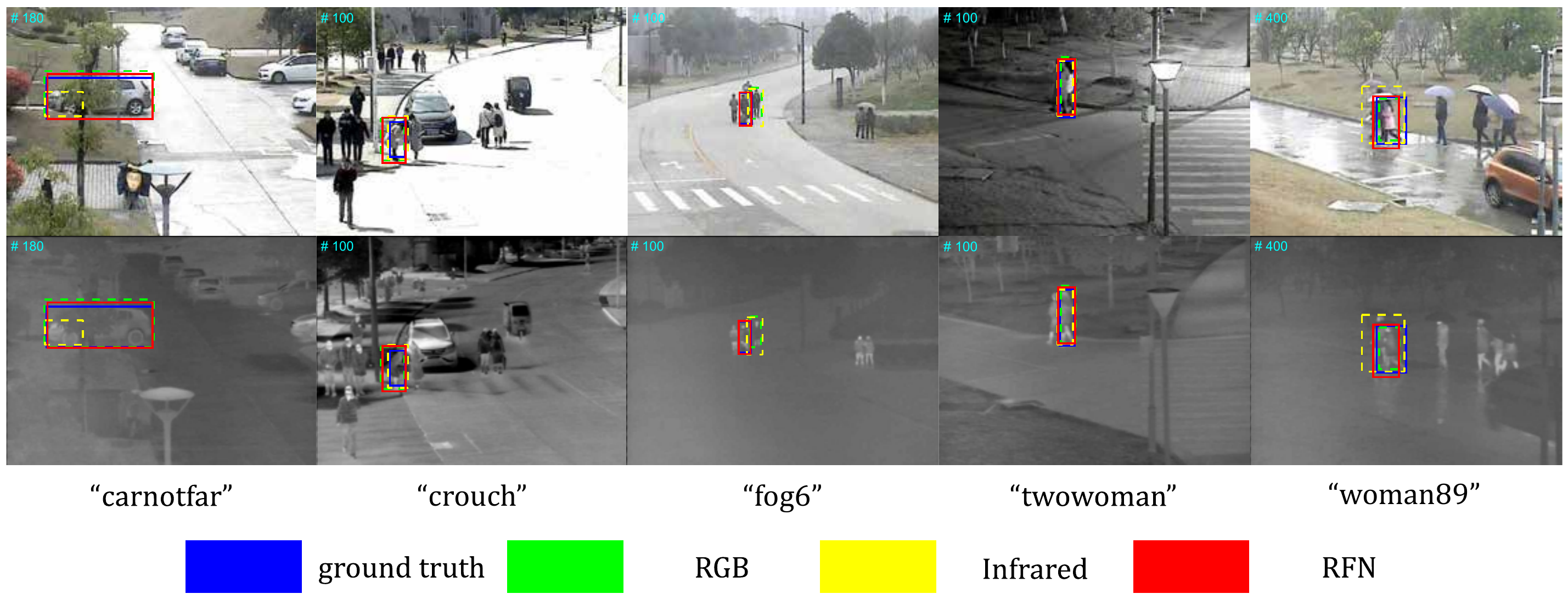}
	\caption{Some tracking results on VOT2020-RGBT. The frames in first row and second row are RGB frames and infrared frames, respectively. Five pairs of RGB and infrared frames in columns are collected from different video sequences (``carnotfar'', ``crouch'', ``fog6'', ``twowoman'' and ``woman89''). AFAT is the base tracker. In these frames, blue boxes denote the ground truth, green boxes and yellow boxes are the tracking results obtained by AFAT with only RGB frames or infrared frames fed as input. Red boxes indicate the tracking results obtained by the RFN-based tracker.}
	\label{fig:tracking}
\end{figure*}

To train the RFN module, the proposed loss function ($L_{RFN}$, Section \ref{sec-train-rfn}) is used in the AFAT training. As RGBT tracking does not involve image generation, it is inevitable that the background detail preservation loss function ($L_{detail}$) needs to be modified to become applicable to the tracking task. $L_{detail}$ is defined as follows,
\begin{eqnarray}\label{equ:tracking-loss-detail}
  	L_{detail}=1 - SSIM(\Phi_{f}^m, l_1(\Phi_{rgb}^m, \Phi_{ir}^m))
\end{eqnarray}
where $\Phi_{f}^m$ denotes the fused deep feature obtained by the RFN module, and $l_1$ indicates the ``$l_1$-norm'' based fusion strategy discussed in Section \ref{sec-abla-rfn}. The target feature enhancement loss function ($L_{feature}$) is the same as in Section \ref{sec-train-rfn}.


\subsection{The Tracking Results on VOT-RGBT}

The video sequences in VOT2020-RGBT are the same as in VOT2019-RGBT. Thus, we only present a few tracking results on VOT2020-RGBT in Fig.\ref{fig:tracking}. The `RFN'  denotes the  RFN-based AFAT.

To evaluate the tracking performance, three measures \citep{vot2019rgbt} were selected: Accuracy ($A$), Robustness ($R$) and Expected Average Overlap ($EAO$). (1) Accuracy denotes the average overlap between the ground truth and the predicted bounding boxes; (2) Robustness evaluates how many times the tracker loses the target (fails) during tracking (3) $EAO$ is an estimator of the average overlap of a tracker. For the detail of Accuracy, Robustness and $EAO$ please refer to \citep{kristan2015visual}.


In VOT2020-RGBT \citep{vot2020rgbt}\footnote{The toolkit version of VOT2020-RGBT is 0.2.0.}, for the Accuracy, Robustness and EAO, they have the same meanings but their calculation methods are re-defined by the committee. Thus, these metrics are indicated as $EAO_{new}$, $A_{new}$ and $R_{new}$. The higher values of $EAO$, $EAO_{new}$, $A$, $A_{new}$, $R$ and $R_{new}$, the better the tracker. 

In addition to the base tracker (AFAT), we choose two further trackers for each dataset (VOT2019, VOT2020) to analyze the tracking performance of RFN-based AFAT. In the VOT2019-RGBT competition, mfDiMP and FSRPN won the third and fourth place on the public dataset, respectively. For VOT2020-RGBT, DFAT and M2C2Frgbt won third and seventh place on the public dataset, respectively. Note that DFAT is the winner on VOT2020-RGBT challenge. All the metrics values are provided by the VOT committee and available on the VOT reports \citep{vot2019rgbt} \citep{vot2020rgbt}.

The tracking results of RFN-based AFAT and other trackers are shown in Table \ref{tab:tracking2019} and Table \ref{tab:tracking2020}. $RGB$ and $Infrared$ indicate that only one modality (RGB or infrared) is fed into AFAT.

\begin{table}[!ht]
	\footnotesize
	\centering
	\caption{\label{tab:tracking2019} The tracking results obtained on the VOT2019-RGBT dataset. AFAT is used as the base tracker to evaluate the RFN network fusion strategy.}
	\resizebox{0.95\linewidth}{!}{
		\begin{tabular}{|c|c|c|c|c|}
			\hline
			\multicolumn{2}{|c|}{\emph{VOT2019}} &$EAO$ &$A$ &$R$ \\
			\hline
			\multicolumn{2}{|c|}{FSRPN\citep{vot2019rgbt}}	&0.3553 	&\emph{\color{red}{0.6362}} 	&\emph{\color{red}{0.7069}} \\
			\hline
			\multicolumn{2}{|c|}{mfDiMP\citep{vot2019rgbt}}	&\textbf{0.3879} 	&0.6019 	&\textbf{0.8036} \\
			\hline
			\multirow{2}*{AFAT\citep{xu2020afat}} &	
			\emph{RGB}		  &0.32590 	&0.61130 	&0.5700 \\
			\cline{2-5}
			&\emph{Infrared}  	&0.18120 	&0.56740 	&0.1800 \\
			\hline
			\multicolumn{2}{|c|}{RFN-based}	&\emph{\color{red}{0.35840}} 	&\textbf{0.64470} 	&0.6500 \\
			\hline
	\end{tabular}}
\end{table}

\begin{table}[!ht]
	\footnotesize
	\centering
	\caption{\label{tab:tracking2020} The tracking results  obtained on the VOT2020-RGBT dataset. AFAT is set as the base tracker to evaluate the RFN network fusion strategy.}
	\resizebox{0.95\linewidth}{!}{
		\begin{tabular}{|c|c|c|c|c|}
			\hline
			\multicolumn{2}{|c|}{\emph{VOT2020}} &$EAO_{new}$ &$A_{new}$ &$R_{new}$ \\
			\hline
			\multicolumn{2}{|c|}{M2C2Frgbt\citep{vot2020rgbt}}	&0.332 	&0.636 	&0.722 \\
			\hline
			\multicolumn{2}{|c|}{DFAT\citep{vot2020rgbt}}	&\textbf{0.390} 	&\textbf{0.672} 	&\textbf{0.779} \\
			\hline
			\multirow{2}*{AFAT\citep{xu2020afat}} &	
			\emph{RGB}		  &0.329 	&0.635 	&0.669 \\
			\cline{2-5}
			&\emph{Infrared}  &0.265 	&0.573 	&0.588 \\
			\hline
			\multicolumn{2}{|c|}{RFN-based}	&\emph{\color{red}{0.371}} 	&\emph{\color{red}{0.668}} 	&\emph{\color{red}{0.726}} \\
			\hline
			
	\end{tabular}}
\end{table}

From these two tables, compared with just feeding one modality into AFAT, the RFN-based AFAT delivers better tracking performance in all measures both on VOT-RGBT2019 and on VOT-RGBT2020. On VOT2019-RGBT, although mfDiMP achieves the best performance, the results produced by the RFN-based tracker are comparable ($EAO$) and the accuracy is better. On VOT2020-RGBT, even compared with the winning tracker, DFAT, our tracker is also competitive. 

These experiments demonstrate that even with insufficient training data, the tracker performance is improved by incorporating the proposed residual fusion network (RFN) into the AFAT tracking framework. When more training data becomes available, we believe the RFN-based tracker will achieve even better tracking performance.

\section{Conclusions}
\label{sec-con}
Motivated by the weakness of the existing fusion methods in preserving image detail, in this paper, we proposed a novel end-to-end fusion framework (RFN-Nest) which is based on the nest connection incorporated into a residual fusion network. To design our RFN-Nest, a two-stage training strategy was presented. In the proposed scheme, an auto-encoder network is trained using the SSIM loss function ($L_{ssim}$) and the pixel loss function ($L_{pixel}$). The trained encoder is utilized to extract multi-scale features from the source images and the nest connection-based decoder network is designed to reconstruct the fused images using the fused multi-scale features. The key component of RFN-Nest is the residual fusion network (RFN). In the second stage of the training strategy, four residual fusion networks (RFN) are trained to preserve the image detail,  and preserve the salient features using $L_{detail}$ and $L_{feature}$, respectively. Once the two-stage training is accomplished, the fused image is reconstructed using the encoder, the RFN networks and the decoder. Compared to seventeen existing fusion methods, the RFN-Nest achieves the best fusion performance in both subjective and objective evaluation.

To validate the generality of the fusion network, we also applied the proposed RFN and the novel loss functions to a state-of-the-art tracker to perform a multimodal tracking task (RGBT tracking). Compared with single modality, the RFN-based tracker delivers better tracking performance in all measures on VOT2019 and VOT2020. Even compared with the state of the art  RGBT trackers, the RFN-based tracker achieves very good performance. This demonstrates that with this proposed innovations, the RFN-Nest network has a wide applicability, extending beyond image fusion.

\section*{Acknowledgement}
This work was supported by the National Natural Science Foundation of China (62020106012, U1836218, 61672265), the 111 Project of Ministry of Education of China (B12018), and the Engineering and Physical Sciences Research Council (EPSRC) (EP/N007743/1, MURI/EPSRC/DSTL, EP/R018456/1).

\bibliographystyle{elsarticle-num-names} 

\bibliography{rfnnest}

\end{document}